\documentclass[acmtog,prologue,table,xcdraw,dvipsnames]{acmart}

\AtBeginDocument{%
  }

\usepackage{booktabs}
\usepackage{multirow}
\usepackage{xspace}

\usepackage{pifont}
\usepackage{tabularx}
\usepackage{graphicx}

\usepackage{siunitx} 
\usepackage{comment}
\usepackage{csquotes}
\usepackage{wrapfig}
\usepackage{cleveref}
\usepackage{color}
\usepackage{lscape}
\usepackage{pifont}

\usepackage{environ}
\usepackage{array}  
\usepackage{multirow}
\usepackage{subcaption}

\usepackage{svg}

\newtoggle{arxiv}

\definecolor{darkred}{RGB}{204, 0, 0}
\definecolor{darkgreen}{RGB}{0, 160, 0}
\definecolor{darkorange}{rgb}{1.0, 0.55, 0.0}
\definecolor{darkorange}{rgb}{1.0, 0.55, 0.0}
\definecolor{darkgreen}{rgb}{0.0, 0.4, 0.0}
\definecolor{darkblue}{RGB}{0, 0, 230}

\newcommand{\greencheck}{\color{PineGreen}\ding{51}}
\newcommand{\orangecheck}{\color{Orange}\ding{51}}
\newcommand{\crossred}{\color{Mahogany}\ding{55}}

\crefname{figure}{Fig.}{Figs.}
\Crefname{figure}{Fig.}{Figs.}

\crefname{table}{Tab.}{Tabs.}
\Crefname{table}{Tab.}{Tabs.}

\crefname{section}{Sec.}{Secs.}
\Crefname{section}{Sec.}{Secs.}

\crefname{subsection}{Sec.}{Secs.}
\Crefname{subsection}{Sec.}{Secs.}

\newcommand{\supmat}[1]{supplementary materials}
\newcommand{\supmatmiddle}[1]{\textcolor{black}{\textit{SupMat}}}
\newcommand{\new}[1]{\textcolor{black}{#1}}

\newcommand{\ie}{\textit{i}.\textit{e}.\,\@\xspace}
\newcommand{\etal}{\textit{et al}.\@\xspace}

\newcommand{\OurMethod}{\textit{3DPR}} 

\newcommand{\OurData}{\textit{FaceOLAT}} 
\newcommand{\OldData}{\textit{WeyrichOLAT}} 
\newcommand{\OurModel}{3DPR_{o}} 
\newcommand{\OldModel}{3DPR_{w}} 

\newcommand{\latentw}{\mathbf{w}}
\newcommand{\dir}{\omega_{i}} 

\newcommand{\egtdfeaturemap}{\mathrm{F_g}}
\newcommand{\olatfeaturemap}{\mathrm{F_o}}

\newcommand{\geomrgb}{\mathrm{c_{rgb}}}
\newcommand{\geomhf}{\mathrm{c_{hf}}}

\newcommand{\olatrgb}{\mathrm{o_{rgb}}}
\newcommand{\olathf}{\mathrm{o_{hf}}}
\newcommand{\fusedolathf}{\mathrm{p_{hf}}}
\newcommand{\imageolat}{\mathbf{O}}
\newcommand{\imagelit}{\mathbf{C}}

\newcommand{\egtdinvnw}{\mathcal{E}}

\newcommand{\egtdgennw}{\mathrm{G_{gen}}} 
\newcommand{\egtddecnw}{\mathrm{G_{dec}}}

\newcommand{\vdir}{\mathbf{v}} 
\newcommand{\olatdecnw}{\mathrm{R_{dec}}}
\newcommand{\olatencodernw}{\mathrm{R_{enc}}} 
\newcommand{\olatsuperresnw}{\mathrm{SR_{o}}}
\newcommand{\olatsuperresencodernw}{\mathrm{E_{SR}}}

\newcommand{\olatencoder}{\emph{Reflectance Encoder}}
\newcommand{\olatdecoder}{\emph{Reflectance Decoder}}
\newcommand{\srencoder}{\emph{SR Encoder}}
\newcommand{\olatsuperres}{\emph{OLAT Super-Resolution}}

\newcommand{\loss}{\mathcal{L}}

\newcommand{\losso}{\loss_\text{O}}

\newcommand{\lossp}{\loss_\text{LPIPS}}
\newcommand{\lossm}{\loss_\text{M}}
\newcommand{\lossmrf}{\loss_\text{MRF}}

\newcommand{\setR}{\mathbb{R}}

\newcommand{\radiance}{\mathrm{O}} 
\newcommand{\incdir}{l} 
\newcommand{\incid}{l} 
\newcommand{\incids}{I} 
\newcommand{\factor}{f} 

\def\rvc{{\mathbf{v}}}

\acmConference[SA Conference Papers '25]{SIGGRAPH Asia 2025 Conference Papers}{December 15--18, 2025}{Hong Kong, Hong Kong}
\acmBooktitle{SIGGRAPH Asia 2025 Conference Papers (SA Conference Papers '25), December 15--18, 2025, Hong Kong, Hong Kong}
\begin{document}
\title{\OurMethod{}: Single Image 3D Portrait Relighting with Generative Priors}

\author{Pramod Rao}
\email{prao@mpi-inf.mpg.de}
\orcid{0009-0003-7236-169X}
\affiliation{%
  \institution{Max Planck Institute for Informatics, SIC \& VIA Research Center}
  \city{Saarbrücken}
  \country{Germany}
}

\author{Abhimitra Meka}
\authornotemark[1]
\email{abhim@google.com}
\orcid{0000-0001-7906-4004}
\affiliation{
  \institution{Google Inc.}
  \city{San Fransisco}
  \country{USA}
}

\author{Xilong Zhou}
\authornote{Equal contribution}
\email{xzhou@mpi-inf.mpg.de}
\orcid{0000-0002-4133-8783}
\affiliation{%
  \institution{Max Planck Institute for Informatics \& SIC }
  \city{Saarbrücken}
  \country{Germany}
}

\author{Gereon Fox}
\email{gfox@mpi-inf.mpg.de}
\orcid{0009-0002-3471-7715}
\affiliation{%
  \institution{Max Planck Institute for Informatics \& SIC }
  \city{Saarbrücken}
  \country{Germany}
}

\author{Mallikarjun B R}
\email{ mbr@mpi-inf.mpg.de}
\orcid{0009-0007-5906-8666}
\affiliation{%
  \institution{Max Planck Institute for Informatics \& SIC }
  \city{Saarbrücken}
  \country{Germany}
}
\author{Fangneng Zhan}
\email{fzhan@mpi-inf.mpg.de}
\orcid{0000-0003-1502-6847}
\affiliation{%
  \institution{Max Planck Institute for Informatics \& SIC }
  \city{Saarbrücken}
  \country{Germany}
}

\author{Tim Weyrich}
\email{tim.weyrich@fau.de}
\orcid{0000-0002-4322-8844}
\affiliation{%
  \institution{Friedrich-Alexander-Universität Erlangen-Nürnberg }
  \city{Nürnberg}
  \country{Germany}
}

\author{Bernd Bickel}
\email{bickelb@ethz.ch}
\orcid{0000-0001-6511-9385}
\affiliation{%
  \institution{ETH Zürich}
  \city{Zürich}
  \country{Switzerland}
}
\author{Hanspeter Pfister}
\email{pfister@g.harvard.edu}
\orcid{0000-0002-3620-2582}
\affiliation{%
  \institution{Harvard University}
  \city{Cambridge}
  \country{USA}
}

\author{Wojciech Matusik}
\email{wojciech@csail.mit.edu}
\orcid{0000-0003-0212-5643}
\affiliation{%
  \institution{Massachusetts Institute of Technology}
  \city{Cambridge}
  \country{USA}
}

\author{Thabo Beeler}
\email{tbeeler@google.com}
\orcid{0000-0002-8077-1205}
\affiliation{
  \institution{Google Inc.}
  \city{Zürich}
  \country{Switzerland}
}

\author{Mohamed Elgharib}
\email{elgharib@mpi-inf.mpg.de}
\orcid{0000-0001-8727-0895}
\affiliation{%
  \institution{Max Planck Institute for Informatics \& SIC }
  \city{Saarbrücken}
  \country{Germany}
}

\author{Marc Habermann}
\email{mhaberma@mpi-inf.mpg.de}
\orcid{0000-0003-3899-7515}
\affiliation{%
  \institution{Max Planck Institute for Informatics, SIC \& VIA Research Center}
  \city{Saarbrücken}
  \country{Germany}
}

\author{Christian Theobalt}
\email{theobalt@mpi-inf.mpg.de}
\orcid{0000-0001-6104-6625}
\affiliation{%
  \institution{Max Planck Institute for Informatics, SIC \& VIA Research Center}
  \city{Saarbrücken}
  \country{Germany}
}

\renewcommand\shortauthors{Rao, P. et al}

%
%
\begin{abstract}
Rendering novel, relit views of a human head, given a monocular portrait image as input, is an inherently underconstrained problem. 
The traditional graphics solution is to explicitly decompose the input image into
geometry, material and lighting via differentiable rendering; but this is constrained by the multiple assumptions and approximations of the underlying models and parameterizations of these scene components.
We propose \OurMethod, an image-based relighting model that leverages generative priors learnt from multi-view One-Light-at-A-Time (OLAT) images captured in a light stage. 
We introduce a new diverse and large-scale multi-view 4K OLAT dataset of 139 subjects to learn a high-quality prior over the distribution of high-frequency face reflectance.
We leverage the latent space of a pre-trained generative head model that provides a rich prior over face geometry learnt from in-the-wild image datasets.
The input portrait is first embedded in the latent manifold of such a model through an encoder-based inversion process. Then a novel triplane-based reflectance network trained on our lightstage data is used to synthesize high-fidelity OLAT images to enable image-based relighting. 
Our reflectance network operates in the latent space of the generative head model, crucially enabling a relatively small number of lightstage images to train the reflectance model. 
Combining the generated OLATs according to a given HDRI environment maps yields physically accurate environmental relighting results.
Through quantitative and qualitative evaluations, we demonstrate that \OurMethod{}  outperforms previous methods, particularly in preserving identity and in capturing lighting effects such as specularities, self-shadows, and subsurface scattering.
\end{abstract}

\begin{teaserfigure}
    \centering
    \includegraphics[width=0.90\linewidth]{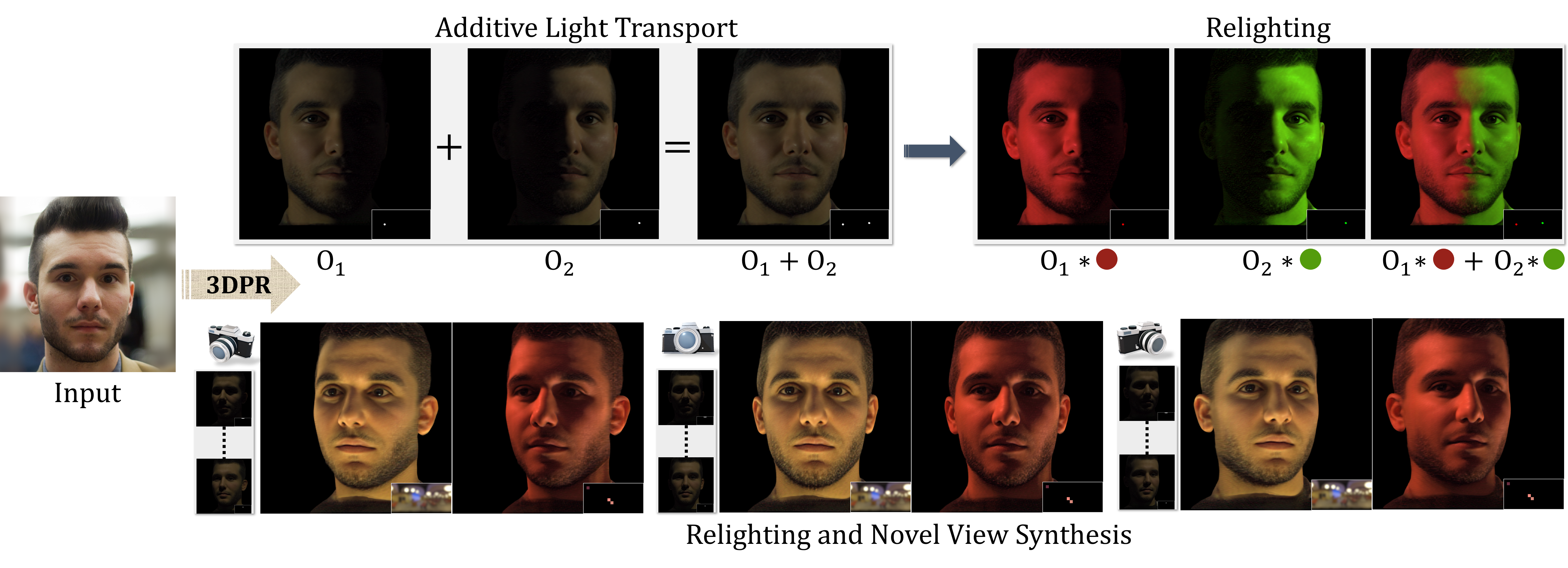}
    \vspace{-0.13in}
    \caption{We present \OurMethod{}, a monocular 3D portrait relighting method that can synthesize novel views under desired illumination.
    Given a monocular input image, \OurMethod{} predicts a reflectance basis in the form of One-Light-At-a-Time (OLAT) images of the subject (top row). By linearly combining the OLAT basis based on given HDRI map, the subject can be placed in novel relit environments (bottom row). Moreover, the OLATs can be rendered for a desired novel camera viewpoint, facilitating 3D-consistent portrait relighting.}
    \label{fig:teaser} 
\end{teaserfigure}

\maketitle
%
%
\section{Introduction} \label{sec:intro}
Modern computer graphics applications, such as Augmented and Virtual Reality, demand blending of real and synthetic assets together seamlessly into a single image. Human faces are of particular importance in such applications and require converting few-shot or even a single monocular face image into 3D assets that can be rendered under novel environments from desired viewpoints to achieve visual immersion.
However, achieving accurate relighting and novel view synthesis in a single unified rendering framework is a highly ill-posed challenge due to the underconstrained problem of 3D modeling from a monocular input and the complexity of the underlying light transport.
%
%
\par
Recently, many data-driven methods have been proposed to learn a 3D prior over the underconstrained solution space of this problem. 
Some methods \cite{gu2021stylenerf,gram,Chan2022} propose generative volumetric representations to synthesize portraits in a 3D-consistent manner. 
Such methods are aimed at solving novel view synthesis and do not necessarily tackle the problem of face relighting. 
More recent methods \cite{deng2023lumigan,neural-3d-relightable, nerffacelighting} build on these volumetric representations and learn a generative model that can also extract reflectance information from a given portrait image.
These methods are trained using
physically-based rendering models and hence suffer from several issues such as the unknown camera and illumination conditions in their in-the-wild training sets. The gap between the underlying physical rendering model and real-world images also significantly affects the quality of photorealism and lighting reproduction. 
To address these issues,  image-based relighting methods \cite{litnerf,yang2023towards,megane_li_cvpr,yang2024vrmm} have been developed using One-Light-at-A-Time (OLAT) \cite{debevec2000acquiring} datasets, which offer ground truth supervision without the need to explicitly model physical light transport.
However, these works  are either subject-specific \cite{litnerf, yang2023towards, saito2024rgca, diffrelight_he} or they require multi-view input during inference \cite{megane_li_cvpr, yang2024vrmm}, and hence do not support generalization to monocular in-the-wild portraits as input. 
\begin{table*}[h!]
\caption{\OurData{} is the first large-scale, publicly available multi-view HDR OLAT face dataset. It includes 139 subjects captured under 3 expressions, illuminated with 331 dense OLAT lighting conditions from 40 viewpoints at 4K resolution. This setup enables high-fidelity full-head reflectance modeling, including hair. None of the existing datasets that are \textit{publicly} available offers this combination of subject diversity, dense illumination, and multi-view coverage at this scale. The {\orangecheck}~symbol for RGCA~\cite{saito2024rgca} indicates the use of grouped OLATs \cite{wenger_2005_sigg}, intended for dynamic capture. ICT-3DRFE~\cite{stratou2011ICTFaceDB} and Ultrastage~\cite{zhou2023relightable} provide only gradient illumination, which is not optimal for high-quality relighting.
}
\vspace{-0.13in}
\begin{center}
\begin{tabular}{|c|c|c|c|c|c|}
\hline
Dataset & \# Illuminations & \# Subjects & \# Views & Resolution & Image-based Relighting \\ \hline
ICT-3DRFE~\cite{stratou2011ICTFaceDB}                     & 3       & 23       & 2           & 1K              & \crossred   \\ \hline
Ultrastage~\cite{zhou2023relightable}                     & 3       & 100      & 32          & 8K              & \crossred   \\ \hline
RGCA~\cite{saito2024rgca, martinez2024codec}              & 460     & 4        & 110         & 4K              & \orangecheck   \\ \hline
Dynamic OLAT~\cite{zhang2021dynamicolat}                  & 114     & 4        & 1           & 1K              & \greencheck  \\ \hline\hline
\OurData                                                  & 331     & 139      & 40          & 4K              & \greencheck  \\ \hline
\end{tabular}
\end{center}
\label{tab:dataset_comparision}
\end{table*}

\par
To relight a single portrait image in a physically-accurate manner, several other works \cite{prao2022vorf, deng2023lumigan, nerffacelighting, mallikarjun2021photoapp, prao2024lite2relight} train a generative model using such an OLAT dataset or GAN-based training\cite{goodfellow2014generative}. 
However, several challenges limit the fidelity of the resulting relighting:
Some such methods \cite{nerffacelighting, deng2023lumigan, mei2024holo, prao2023vorf} require test-time optimization for each subject, which is very time-consuming and impractical for most AR/VR applications.
Other methods \cite{prao2024lite2relight, mallikarjun2021photoapp} propose lightweight solutions, 
that utilize 2D generative models like StyleGAN\cite{Karras2019stylegan2} or EG3D \cite{Chan2022} that restricts detailed face reflectance modelling and struggle with complex lighting effects, such as accurate shadows and specularities (\cref{fig:baseline_nfl_l2r-wild_cts_env_same_view}).
VoRF \cite{prao2022vorf}, also a volumetric relighting technique, excels at capturing shadows as it models face reflectance through a set of OLAT basis functions. 
However, its results are over-smooth and it struggles to generalize to in-the-wild portraits as it is trained on a  limited set of face OLAT images that do not capture a rich geometry and appearance prior.
No large-scale light stage dataset of sufficient diversity is currently publicly available.
\par 
To address these challenges, we present \OurMethod{}, a 3D portrait relighting method that leverages volumetric generative models combined with a novel light stage dataset. We introduce \OurData{}, a large-scale, high-quality, human face OLAT dataset that will be publicly released for the benefit of the research community. 
\OurMethod{} takes a monocular portrait as input and renders 3D-consistent novel views under any given lighting, accurately simulating complex light transport effects (\cref{fig:teaser}).
To achieve this, our method synthesizes OLAT images for the desired viewpoint, that are then linearly combined according to an HDRI map.  
We build on EG3D \cite{Chan2022}, which provides a robust generative prior for facial geometry and appearance, enabling our approach to effectively generalize to unseen faces. 
The input portrait is embedded into EG3D's latent space via encoder-based GAN inversion  \cite{goae}. 
Crucially, we combine EG3D with our novel reflectance model, efficiently encoding face reflectance into a triplane representation, which allows rendering high-resolution OLAT images.
\par
Our dataset, \OurData{}, offers 40 camera viewpoints at 4K resolution and 331 point light sources, surpassing all publicly available datasets (see \cref{tab:dataset_comparision}) as well as the \textit{non-public} OLAT dataset of \citet{Weyrich2006Analysis}, which is widely used for evaluating face reflectance modelling.
\OurData{} includes subjects of different skin tones, hair colors, eye colors, ethnicities and ages, providing demographic diversity. 
Training \OurMethod{} on \OurData{} leads to  state-of-the-art results, both quantitatively and qualitatively. 
\par 
In summary, we contribute:
%
\begin{itemize}
    \item An image-based 3D portrait relighting method leveraging a combination of pretrained generative prior and an OLAT dataset to enable physically accurate editing of both illumination and viewpoint of a monocular input image. 
    \item  \OurData{}, a large face OLAT dataset, comprising 139 subjects captured by 40 cameras under 331 point light sources. Our dataset will be publicly available.
\end{itemize}
Comprehensive quantitative and qualitative evaluations shows that our method achieves state-of-the-art performance.
Code and pre-trained checkpoints is available under 
\textcolor{CarnationPink}{\url{https://vcai.mpi-inf.mpg.de/projects/3dpr/}}.
\section{Related Work}

Many portrait relighting methods employ illumination models that are trained on synthetic data~\cite{Shu17NeuralEditing,sfsnetSengupta18,Zhou_2019_ICCV,Chandran_2022_WACV,lattas2021avatarme++}. While these methods do generalize to novel identities, their photorealism and overall quality leave room for improvement \cite{Shu17NeuralEditing,sfsnetSengupta18,Zhou_2019_ICCV}. Modeling the complex light transport effects exhibited by human faces, as well as the sub-surface material properties of skin \cite{klehm15star} is a challenging task.
This is why a different line of research,
image-based relighting \cite{debevec2000acquiring} uses OLAT images captured with a light stage as a basis for relighting according to any given HDRI environment map. 
This approach has been extended to estimating reflectance  from monocular images \cite{mbr_frf,yamaguchi18}, based on  parametric face models:
FRF \cite{mbr_frf} aims to regress an OLAT basis for a given camera view but the parametric face model limits it to the face interior.
Similarly, many methods are limited to portrait relighting without view synthesis \cite{zhang2020portrait,Meka19,sun2020light,nestmeyer2020faceRelighting,Pandey21,zhang2021dynamicolat, zhang2025_ic_light, zeng2024_dilightnet} or subject-specific relighting \cite{Bi21}. 

Some methods learn face priors using 2D generative models, adapting them for photorealistic editing of pose, expression, and lighting \cite{stylerig, tewari2020pie, mallikarjun2021photoapp, buehler2021varitex}. 
Specifically, PhotoApp \cite{mallikarjun2021photoapp} combines the advantages of lightstage OLAT data and a generative StyleGAN model, resulting in impressive identity generalization, simultaneous relighting and novel view synthesis of the full head. Nevertheless it suffers from view inconsistency and fails to preserve the original identity due to the absence of a consistent 3D facial geometry representation. 
In contrast, our method leverages a prior in volumetric space. This results in improved view consistency and ensures the preservation of the original identity throughout the editing process.

Neural field techniques have achieved high photorealism in view synthesis, but relighting remains an open problem. Srinivasan \etal \cite{Srinivasan21NeRV} show relighting in general scenes using co-located camera and light source, for a dense set of input images. More recent work \cite{Zhang21,Boss20NeRD,rudnev2022nerfosr} has extended this to images captured under unknown lighting, but these methods are scene-specific and cannot generalize to monocular inputs for an object category like faces or heads. 
Hong \etal \cite{headnerf} build a parametric head model conditioned on a lighting latent code. They disentangle lighting and reflectance by supervision on a multi-light dataset, but are limited by the sparse lighting variation in the training data. Kwak \etal \cite{kwak2022injecting} attempt to decouple semantic attributes (including lighting) but suffer from significant view inconsistency due to the underlying unsupervised training scheme. Other methods \cite{sun2021nelf,prao2022vorf} achieve view synthesis and relighting of real people from sparse images:
NeLF \cite{sun2021nelf} relies on a pixelNeRF-inspired \cite{yu2020pixelnerf} architecture and thus struggles to capture global features.
Holo-relighting \cite{mei2024holo} also leverages an EG3D prior to disentangle, delight and then relight a volumetric face from a single image, using lightstage data. However, it does not estimate  intermediate OLAT images, but relies on neural networks to fully interpret an environment map, giving more opportunity for physically implausible results. It also requires test-time optimization which is computationally expensive and time consuming.

Some approaches \cite{tan2022voluxgan, neural-3d-relightable} focus on relighting synthetic identities sampled from a learned latent space, but cannot relight a given real image.
In contrast, LumiGan \cite{deng2023lumigan} can very well relight a given image, but while its adversarial self-supervised training leads to plausible-looking outputs, it does not  supervise actual physical accuracy.

 Both Lite2Relight \cite{prao2024lite2relight} and NeRFFaceLighting \cite{nerffacelighting} use a triplane representation  \cite{Chan2022}. While the former trains on a light stage dataset and produces physically accurate relighting, the latter trains on an in-the-wild dataset. However, NeRFFaceLighting uses spherical harmonics (SH), restricting its results to low-resolution lighting conditions and Lite2Relight samples the target lighting from the latent space of the 3D generator. In contrast, our method explicitly synthesizes OLAT images, which are then linearly combined according to an HDRI map. 

The method that overcomes most of the aforementioned challenges is VoRF
\cite{prao2022vorf, prao2023vorf}. It is the closest related work in terms of problem setting. VoRF builds on a light stage dataset to learn physically accurate lighting.
However, VoRF's face prior, learned from a relatively small number of light stage subjects, struggles to generalize  to monocular inputs of unseen faces. Our method addresses this problem by combining a generative 3D face prior \cite{Chan2022} with light transport learned from a lightstage dataset. This leads to 3D-consistent novel-view synthesis and physically accurate relighting. 

\begin{figure*}
    \centering
    \includegraphics[width=\linewidth]{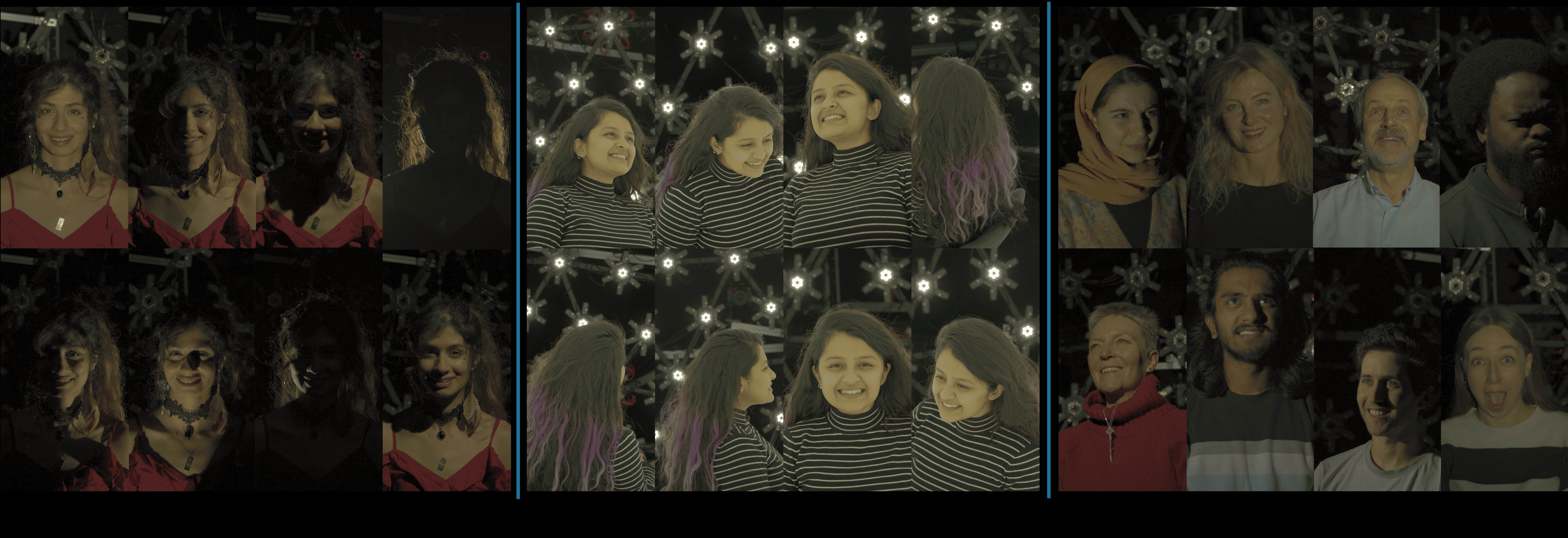}
    \caption{\textbf{Overview of the Dataset.}
    The \OurData{} lightstage dataset comprises 139 subjects captured from 40 camera viewpoints, resulting in 331 OLAT images per subject, illuminated by point light sources. Each OLAT image is captured at 4K resolution. A snapshot of the dataset is shown in the figure.
    A detailed description along with dataset demographics is provided in the supplemental material.
    }\label{fig:dataset_overview}
\end{figure*}

\section{\OurData{}: A New Large-Scale OLAT Dataset}
\label{sec:dataset}
Image-based relighting methods benefit from directly leveraging captured reflectance data without relying on any predefined material models.
They achieve this by linearly combining One-Light-At-a-Time (OLAT) images.
Given an HDR environment map specifying illumination, the relit image $\imagelit$ is computed as: $\imagelit \approx \sum_{\incid \in \incids} \factor_\incid \cdot \radiance(\incid)$, where $\incids$ denotes the set of OLAT lighting directions, $\radiance(\incid)$ is the OLAT image for lighting from direction $\incid$, and $\factor_\incid$ represents the environment map weights.

To effectively train our model to synthesize OLAT images (\cref{sec:methodology}), a high-quality multi-view OLAT dataset is essential. However, existing publicly available datasets are limited in scale and diversity \cite{zhang2021dynamicolat, saito2024rgca}. Addressing this significant gap, we introduce \OurData, a novel, large-scale OLAT dataset comprising 139 diverse subjects captured using a well-calibrated lightstage system. Each subject was recorded from 40 uniformly distributed viewpoints at 4K resolution under 331 OLAT lighting conditions, capturing four distinct facial expressions. The \cref{fig:dataset_overview} provides an overview of the dataset. 
Our detailed dataset capture pipeline resolves practical challenges, such as minor involuntary subject movements during the \num{7}\si{\second} capture duration, by interleaving fully lit reference frames every 21 OLAT captures and employing optical flow-based alignment \cite{Teed2020RAFT}.
Additional preprocessing includes precise calibration, detailed 3D reconstruction, and efficient background segmentation using BGMv2 and RMBGv2 \cite{BGMv2,rmbgv2_BiRefNet}. We partition the dataset into training and evaluation subsets, with 129 subjects designated for training and the remaining 10 for evaluation.
Our dataset, which includes the preprocessing results, such as 3D reconstructions, will be publicly accessible. Additionally, the supplemental document contains additional information on demographics, preprocessing techniques, and dataset acquisition. We now detail our proposed methodology that leverages this dataset
%

%
\begin{figure*}[t]
    \centering
    \includegraphics[width=\linewidth]{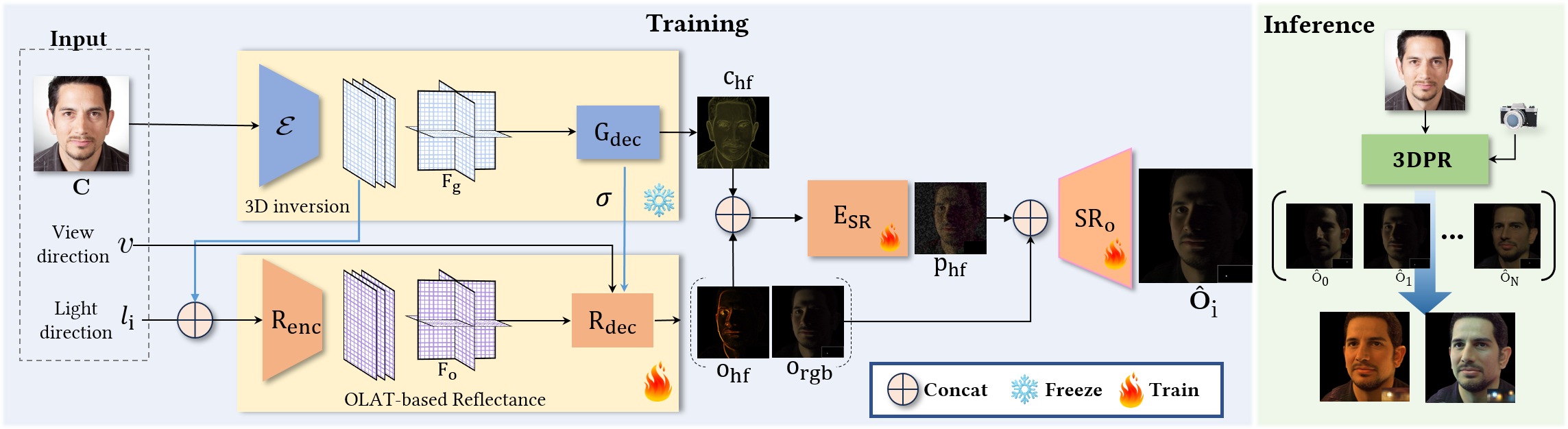}
    \vspace{-0.13in}
    \caption{ 
    Given one portrait image $\imagelit$, \OurMethod{} renders the subject from a novel viewpoint under new lighting. \textbf{Left}: During the training stage, $\imagelit$ is first fed into the 3D-aware encoder $\egtdinvnw$~\cite{goae} to produce tri-planar features $\egtdfeaturemap$, concatenated with a given light direction $\incdir_i$. Then, the concatenated features are fed into the \olatencoder{} $\olatencodernw$ and \olatdecoder{} $\olatdecnw$ to render a low-resolution OLAT image $\olatrgb$ and high-frequency reflectance features $\olathf$. We further combine $\olathf$ with the corresponding appearance features $\geomhf$, to be fed into \srencoder{} $\olatsuperresencodernw$ to obtain fused high frequency features $\fusedolathf$. At the end, the \olatsuperres{} network $\olatsuperresnw$ produces a high-resolution OLAT image $\hat\imageolat$. In our architecture, we use pre-trained networks~\cite{Chan2022} for StyleGAN and $\egtddecnw$ and keep these modules frozen, and only train $\olatencodernw$, $\olatdecnw$, $\olatsuperresencodernw$ and $\olatsuperresnw$ on light stage dataset. \textbf{Right}: During inference, \OurMethod{} takes a single portrait image, view and light direction as input, and synthesize OLATs, which are then linear combined for novel illumination.
    }\label{fig:pipeline_training}
\end{figure*}

\section{Method} \label{sec:methodology}
Given a single portrait image, our goal is to edit both viewpoint and illumination in a photorealistic and 3D-consistent manner. To achieve this, \OurMethod{} is trained on an OLAT dataset and operates in two stages:
In the first stage, the input portrait is embedded into the latent space of EG3D via an encoder-based GAN inversion process (see \cref{subsec:inversion}), enabling our framework to benefit from EG3D’s strong 3D generative prior.
In the second stage, we introduce an OLAT-based reflectance module that synthesizes OLAT images using an efficient volumetric triplane representation~\cite{Chan2022}; this stage is described in detail in \cref{subsec:reflectance}.
During training, the reflectance module is supervised using ground-truth OLAT images from the lighstage dataset (see \cref{subsec:loss_functions} and supplemental for additional training details).
At inference time, \OurMethod{} takes a single RGB input image and synthesizes OLAT images for novel viewpoints, which are then linearly combined to approximate the target lighting condition (see \cref{subsec:testing}).
\subsection{3D Inversion} \label{subsec:inversion}
EG3D transforms a noise vector $\mathbf{z} \in \setR^{1 \times 512}$ into an intermediate latent code $\latentw \in \setR^{14 \times 512}$, which is passed to a StyleGAN2 generator $\egtdgennw$~\cite{Karras2019stylegan2} to produce tri-planar features $\egtdfeaturemap \in \setR^{96 \times 256 \times 256}$. These features encode both geometry and appearance, and serve as a compact 3D representation of the scene. They can be rendered to images from arbitrary viewpoints, by volume rendering.
To obtain this feature representation from a single portrait image $\imagelit$, we employ a pre-trained encoder-based inversion network $\egtdinvnw$~\cite{goae}, such that $\egtdfeaturemap = \egtdinvnw(\imagelit)$ (see \cref{fig:pipeline_training}).
The tri-plane features $\egtdfeaturemap$ are decoded by EG3D’s MLP-based decoder $\egtddecnw$ and rendered volumetrically from a given camera viewpoint $\rvc$ to produce a low-resolution RGB image $\geomrgb \in \setR^{3 \times 128 \times 128}$ and a high-frequency feature image $\geomhf \in \setR^{29 \times 128 \times 128}$. 
We specifically leverage $\geomhf$, which encodes high-frequency appearance details, as input to the next stage of our pipeline for synthesizing high-resolution OLATs.
%
%
%
\subsection{Learning Face Reflectance} \label{subsec:reflectance}
To model facial reflectance, we aim to generate OLAT images for any light direction $\dir \in \setR^3$. Given the encoded tri-plane feature map $\egtdfeaturemap$ from the inversion stage, we concatenate it with the light direction $\dir$ and pass the result to our OLAT encoder $\olatencodernw$, which predicts reflectance-aware tri-plane features $\olatfeaturemap \in \setR^{96 \times 256 \times 256}$ as:
$\olatfeaturemap = \olatencodernw(\egtdfeaturemap, \dir)$.
$\olatencodernw$ is based on a ResNet architecture~\cite{he2016residual}, and the depth of 96 channels is critical for modeling complex skin-light interactions such as specularities, hard shadows, and subsurface scattering.

To synthesize a specific OLAT image from these features, we also incorporate the view direction $\vdir$ and use our OLAT decoder $\olatdecnw$, a lightweight single-layer MLP, which outputs both a low-resolution RGB image $\olatrgb \in \setR^{3 \times 128 \times 128}$ and a high-frequency reflectance feature map $\olathf \in \setR^{29 \times 128 \times 128}$ via NeRF-style volume rendering:
$\olatrgb, \olathf = \olatdecnw(\olatfeaturemap, \vdir)$. 
%
If we directly fed $\olatrgb$ and $\olathf$ into the super-resolution module $\olatsuperresnw$, that module could easily overfit to the realitvely small number of subjects in the OLAT dataset.
To prevent this, we introduce a feature fusion module $\olatsuperresencodernw$, which combines $\olathf$ with the high-frequency identity features $\geomhf$ obtained from the inversion stage (see \cref{subsec:inversion}). The fused feature map $\fusedolathf \in \setR^{29 \times 128 \times 128}$ is computed as:
$\fusedolathf = \olatsuperresencodernw(\geomhf \oplus \olathf)$,
where $\oplus$ denotes channel-wise concatenation.
Since $\olatsuperresnw$ was pretrained to use $\geomhf$, the module $\olatsuperresencodernw$ quickly learns to forward much of the information from $\geomhf$ to $\olatsuperresnw$. Only for lighting information is $\olatsuperresnw$ forced to rely on $\olatrgb$ and $\olathf$. Information about the identity of the subject however, is contained in $\geomhf$ from the start of training, which acts as a kind of regularization that prevents $\olatsuperresnw$ from trying to derive the identity from  $\olatrgb$ and $\olathf$. Our experiments show (see \cref{subsec:ablations,tab:ablation-losses-sr_enc}) that introducing $\olatsuperresencodernw$ does improve quality. 
%
%
Finally, we combine $\olatrgb$ and $\fusedolathf$ and pass them through the OLAT super-resolution network $\olatsuperresnw$ to synthesize the final high-resolution OLAT image 
$\hat{\imageolat} = \olatsuperresnw(\olatrgb \oplus \fusedolathf) \in \setR^{3 \times 512 \times 512}$.
This pipeline enables accurate and generalizable OLAT image synthesis, which is used for relighting under arbitrary environment maps.

\subsection{Loss Functions} \label{subsec:loss_functions}
\paragraph{Reconstruction Loss.}
Given ground-truth OLAT images $\imageolat_i$ from the light stage, we supervise our predicted OLATs $\hat{\imageolat}_i$ using the $L1$ loss $\losso = \lVert \hat{\imageolat}_i - \imageolat_i \rVert_1$. This encourages direct correspondence between the predicted and reference images at the pixel level.

\paragraph{ID-MRF Loss.}
Relying solely on $\losso$ is insufficient, as misalignments introduced by the inversion stage lead to subtle errors that cannot be corrected by pixel-wise losses. Adversarial losses are also unsuitable here, given the limited number of subjects in the dataset (130), which risks discriminator overfitting and unstable training.
To address this, we adopt the Implicit Diversified Markov Random Field (ID-MRF) loss introduced by Wang \etal~\cite{wang2018mrfloss}. This loss encourages local feature-level similarity by minimizing the patch-wise nearest-neighbor distances between $\hat{\imageolat}_i$ and $\imageolat_i$ in a feature space extracted from a pre-trained VGG19 network~\cite{Simonyan15}.
The ID-MRF loss is computed as
$\lossmrf = \lossm(\Phi_1(\hat{\imageolat}_i, \imageolat_i)) + \lossm(\Phi_2(\hat{\imageolat}_i, \imageolat_i))$,
where $\Phi_1$ and $\Phi_2$ correspond to activations from the \texttt{conv3\_2} and \texttt{conv4\_2} layers of VGG19, respectively, and $\lossm$ denotes the matching function.

\paragraph{Final Objective.}
The total loss is given by $\loss = \losso + 0.3 \lossmrf$, where the ID-MRF term is weighted to balance reconstruction accuracy with local structural detail. As shown in \cref{subsec:ablations}, this loss formulation recovers high-frequency details more effectively than commonly used perceptual losses such as LPIPS~\cite{zhang2018perceptual}.

%
%
%
%
\subsection{Testing} \label{subsec:testing}
At test time, given a monocular portrait image, we employ the encoder-based inversion network to derive $\egtdfeaturemap$. 
This feature map is processed through the reflectance network, which synthesizes OLAT images corresponding to a specified lighting direction and camera viewpoint.
The design of the 3D generative model enables the generation of OLAT images in a single forward pass, eliminating the need for computationally intensive test-time optimization processes like in VoRF \cite{prao2023vorf, prao2022vorf} or NFL \cite{nerffacelighting}. 
%
Finally, exploiting the additivity of light transport, the predicted OLAT images can be linearly combined with the desired HDR environment maps (see \cref{sec:dataset}). This allows relighting of the portrait under the desired illumination conditions while supporting novel viewpoints. Our inference pipeline is visualized in \cref{fig:pipeline_training}.
\input{figures/tex_figures/ours_in_the_wild_01}

\section{Results and Discussion}\label{sec:Results}

We evaluate \OurMethod{} on two categories of datasets: For qualitative evaluation, we assess simultaneous view synthesis and relighting on in-the-wild subjects from RAVDESS~\cite{ravdess}, Flickr~\cite{shih2014style}, and FFHQ~\cite{Karras2019stylegan2}. For quantitative evaluation, we use $\OldData$~\cite{Weyrich2006Analysis} and \OurData{}, where we construct input-reference image pairs by selecting 10 unseen subjects and relighting them using 10 novel HDR environment maps (\cref{sec:dataset}).
\cref{subsec:relighting_wild} presents qualitative results of simultaneous relighting and view synthesis on in-the-wild data using \OurMethod. We further compare our approach both qualitatively and quantitatively to state-of-the-art methods in \cref{subsec:baseline_comparison}. Finally, we analyze key design choices through ablation studies in \cref{subsec:ablations}.

\subsection{Qualitative Evaluation}
\label{subsec:relighting_wild}
\cref{fig:ours_in_the_wild_1} presents qualitative results for simultaneous view synthesis and relighting. \OurMethod{} preserves the linear nature of light transport and accurately reproduces illumination effects such as hard shadows, specular highlights, and self-shadowing, all consistent with the reference images. For instance, observe the shading details on the nose and cheek regions in the OLAT renderings (2nd to 4th columns).
Our method leverages a rich 3D generative prior and models facial reflectance through OLAT-based reflectance module, enabling it to capture the complex interplay between light, face geometry and skin. This design allows \OurMethod{} to faithfully relight subjects while preserving facial structure and expressions from the input.
Overall, our qualitative results show that \OurMethod{} produces accurate relighting that is 3D-consistent across diverse subjects.

\subsection{Baseline Comparisons}
\label{subsec:baseline_comparison}
We compare \OurMethod{} against several state-of-the-art approaches for simultaneous view synthesis and relighting:
\begin{itemize}
  \item \textbf{PhotoApp}~\cite{mallikarjun2021photoapp} leverages the generative prior of StyleGAN2~\cite{Karras2019stylegan2} to learn a latent space transformation for portrait relighting.
  \item \textbf{VoRF}~\cite{prao2022vorf, prao2023vorf} trains an autodecoder-based NeRF~\cite{mildenhall2020nerf} to learn a volumetric reflectance field of human heads.
  \item \textbf{NeRFFaceLighting (NFL)}~\cite{nerffacelighting} disentangles lighting and appearance using EG3D-based design principles and performs relighting via an SH-based representation.
  \item \textbf{Lite2Relight (L2R)}~\cite{prao2024l2r} employs an MLP-based reflectance network that probes the latent space of EG3D to enable controllable relighting.
\end{itemize}



For a fair comparison, we evaluate all methods on \OldData{}, a well-established (non-open-source) benchmark, using the same train–test split as L2R to ensure standardized evaluation. 
\new{In \cref{tab:baseline-LS_weyrich}, all baselines and our method ($\OldModel$) are trained and evaluated on \OldData{}. Thus, the observed improvements of \OurMethod{} arise from the effective way of combining 3D generative priors with OLAT representation. Further, in \cref{tab:baseline-LS_ours}, we benchmark our method against the strongest baselines on \OurData{} and retrain L2R on our dataset to ensure a fair comparison.}

We measure relighting accuracy using multiple metrics: LPIPS~\cite{zhang2018perceptual}, RMSE, DISTS~\cite{dists_metric}, PSNR, SSIM, and identity consistency (ID), computed as the cosine similarity between MagFace~\cite{meng2021magface_ID_loss} embeddings of the relit and ground-truth images.

Quantitative comparisons with all baselines on \OldData{} are shown in \cref{tab:baseline-LS_weyrich} and \cref{fig:baseline-LS}. To further validate generalization and performance, we also evaluate \OurMethod{} against the two strongest baselines, NFL and L2R, on our new \OurData{} dataset; results are presented in \cref{tab:baseline-LS_ours} and \cref{fig:baseline_nfl_l2r-wild_cts_env_same_view}. For this evaluation, we retrain L2R following its original training protocol.

\begin{table}[h!]
    \centering
\caption{\textbf{Quantitative Comparisons:  NeLF~\cite{sun2021nelf}, PhotoApp~\cite{mallikarjun2021photoapp}, VoRF~\cite{prao2022vorf, prao2023vorf}, NeRFFaceLighting~\cite{nerffacelighting} and Lite2Relight~\cite{prao2024l2r}}. Metrics evaluated on the \OldData{} test set, for simultaneous view synthesis and relighting.} 
        \vspace{-0.13in}
        \label{tab:baseline-LS_weyrich}
        \centering
        \resizebox{\columnwidth}{!}{%
        \begin{tabular}{lccccccc}
        \hline
        \textbf{}        & SSIM$\uparrow$   & LPIPS$\downarrow$  & RMSE$\downarrow$         & DISTS$\downarrow$        & PSNR$\uparrow$ & ID$\uparrow$        \\ \hline
        NeLF             & 0.75          & 0.4874           & 0.2466          & 0.2212          & 19.72          & 0.798 \\ 
        PhotoApp         & 0.72          & 0.4163           & 0.1988          & 0.2031          & \textbf{29.13} & 0.853 \\ 
        VoRF             & 0.69          & 0.3253           & 0.1967          & 0.1934          & 20.21          & 0.860 \\ 
        NeRFFaceLighting & 0.79          & 0.2171           & 0.2393          & 0.2107          & 27.24          & 0.892 \\ 
        Lite2Relight     & 0.83          & 0.2492           & 0.1841          & 0.1719          & 28.27          & 0.936 \\
        \textbf{Ours ($\OldModel$)}    & \textbf{0.87} & \textbf{0.1828}  & \textbf{0.1332} & \textbf{0.1689} & 28.69   & \textbf{0.942}       \\ \hline
        \end{tabular}
        }
\end{table}
\begin{table}[h!]
    \centering
        \caption{\textbf{Quantitative Comparisons:  NeRFFaceLighting~\cite{nerffacelighting}, Lite2Relight~\cite{prao2024l2r}}. Performance metrics are evaluated on the \OurData{} test dataset, for simultaneous view synthesis and relighting.}
        \vspace{-0.13in}
        \label{tab:baseline-LS_ours}
        \centering
        \resizebox{\columnwidth}{!}{%
        \begin{tabular}{lccccccc}
        \hline
        \textbf{}        & SSIM$\uparrow$   & LPIPS$\downarrow$  & RMSE$\downarrow$         & DISTS$\downarrow$        & PSNR$\uparrow$ & ID$\uparrow$         \\ \hline
        NeRFFaceLighting & 0.77          & 0.2385           & 0.2926          & 0.2193          & 16.97     & 0.906     \\ 
        Lite2Relight     & 0.79          & 0.2506           & 0.2619          & 0.20861          & 16.72    & 0.910   \\ 
        \textbf{Ours ($\OurModel$)}    & \textbf{0.83} & \textbf{0.1996}  & \textbf{0.1801} & \textbf{0.1751} & \textbf{21.02} & \textbf{0.943} \\ \hline
        \end{tabular}
        }
\end{table}

\input{figures/tex_figures/baseline_comparison}

\paragraph{Relighting and Novel View Synthesis.}
\cref{tab:baseline-LS_weyrich,tab:baseline-LS_ours} report quantitative comparisons, while \cref{fig:baseline-LS} shows qualitative examples. These results confirm that \OurMethod{} outperforms competing methods both numerically and visually.
PhotoApp lacks an explicit 3D representation, often resulting in identity inconsistencies (e.g., altered jawlines) under novel viewpoint and fails to capture accurate illumination effects. Despite this, it achieves surprisingly high PSNR scores, largely due to the high visual quality of StyleGAN2-generated images. However, PSNR is not well-suited for measuring the variations in complex illuminations and thus cannot properly account for the nuances of human visual perception \cite{zhang2018perceptual}.
\input{figures/tex_figures/baseline_new_SOTA_input_view_cts}
VoRF struggles with accurate OLAT synthesis due to inference-time optimization that modifies the learned volumetric reflectance representation. Furthermore, its limited face prior restricts its ability to generalize to unseen identities.
NFL also suffers from challenges in relighting accuracy and identity preservation, mainly due to its two-stage optimization pipeline. As visualized in \cref{fig:baseline-LS}, input lighting is often baked into the albedo, and NFL fails to capture high-frequency details.
Finally, L2R relights portraits by predicting latent vectors through probing EG3D's manifold, but lacks precise lighting control. As shown in \cref{fig:baseline-LS}, it fails to reproduce soft shadows and yields inconsistent facial illumination relative to the ground truth.
In contrast, \OurMethod{} leverages the additive nature of light transport, enabling fine-grained lighting control while faithfully preserving identity -- even under novel viewpoints. This demonstrates the method’s robustness in modeling complex light interactions and its ability to maintain photorealism and consistency across a wide range of subjects and conditions. Please refer to \supmat{} for additional comparisons.

\paragraph{Significance of OLAT-based Relighting}
Both NFL and L2R leverage the EG3D generative prior to directly predict relit portraits. In contrast, our approach not only incorporates the EG3D prior but also explicitly models facial reflectance via OLAT prediction and linear combination with environment maps. This design offers fine-grained control over lighting and enables relighting under \textit{any} lighting condition, including artistic, sparse, or non-natural illumination setups commonly used in cinematic and indoor environments.
We hypothesize that such conditions fall outside the training distribution of EG3D, which is primarily trained on in-the-wild images with natural illuminations. To evaluate this, we create increasingly sparse lighting conditions by randomly replacing environment map pixels with zero (see the first row in \cref{fig:baseline_nfl_l2r-wild_cts_env_same_view}). As the lighting becomes sparser (left to right), both NFL and L2R exhibit noticeable performance degradation, confirming our hypothesis.
While L2R performs reasonably under dense lighting, it fails under sparse setups due to out-of-distribution target illuminations. NFL, limited by its SH-based lighting representation and inaccurate albedo-lighting disentanglement, struggles to reproduce high-frequency effects and breaks down under colored light. In contrast, \OurMethod{} remains robust across lighting conditions, accurately reproducing shadows, specularities, and other complex effects even under sparse or unconventional illumination.
See the supplementary material and caption of \cref{fig:baseline_nfl_l2r-wild_cts_env_same_view} for detailed analysis and visual examples.

\paragraph{Quality of OLATs}
We quantitatively evaluate the accuracy of OLAT renderings produced by \OurMethod{}, obtaining significantly improved results (SSIM: $0.88$, LPIPS: $0.1753$, PSNR: $28.70$) compared to the state-of-the-art method VoRF (SSIM: $0.71$, LPIPS: $0.3148$, PSNR: $20.43$). Qualitative results in \cref{fig:ablation-olats-wild} demonstrate that our synthesized OLATs generalize robustly to both our evaluation dataset and \enquote{in-the-wild} subjects. Our method effectively preserves the additive properties of light transport, and accurately reproduces complex illumination effects, including specular highlights, hard shadows, and subsurface scattering effects. 

\input{figures/tex_figures/ablation_olats}
\subsection{Ablation Study}  
\label{subsec:ablations}  

\paragraph{Timing Evaluations:}
On an NVIDIA 3090 GPU, \OurMethod{} synthesizes the complete set of 331 OLAT images in approximately 30.49 s. We observe that this number can be reduced to 150 OLATs with minimal degradation in quality, reducing the runtime to around 13.8 s. Since OLAT synthesis is fully parallelizable, using an H100 GPU reduces the time for generating all 331 OLATs to just 7.74 s.
Importantly, in 3DPR, OLATs for a given subject and viewpoint are rendered only once. Once these are generated, relighting under a novel environment map takes just 0.24 s. Although our method is not as fast as Lite2Relight, we believe it offers a practical balance between efficiency and quality. Compared to optimization-based baselines, it is competitively fast and significantly outperforms all baselines, including Lite2Relight, in relighting fidelity.
Furthermore, because our approach models a continuous reflectance field, it naturally supports flexible upsampling of lighting resolution. For instance, we can synthesize 1324 OLAT images in just 34.64 s on a H100 GPU, demonstrating the scalability of our method.

\paragraph{Significance of SR Encoder:}  
Given the relatively small size of the lightstage training dataset with its limited subject diversity, the reflectance module, particularly the $\olatsuperresnw$ network, tends to overfit during relighting. 
This overfitting is evident in artifacts observed on subjects not included in the lightstage dataset (refer to \supmat{}).  
To address this issue and improve generalization, we combine robust high-frequency face prior features  $\geomhf$  with high-frequency reflectance features  $\olathf$  using $\olatsuperresencodernw$. This integration mitigates the memorization of training subjects and enhances overall performance, as summarized in \cref{tab:ablation-losses-sr_enc}.  

\paragraph{Significance of $\lossmrf$:}  
\begin{table}[t]
\caption{\textbf{Quantitative Results: Design Ablations}: We report the influence of various losses and $ \olatsuperresencodernw$. The performance metrics  are evaluated for relighting performance across 10 unseen subjects \cite{Weyrich2006Analysis}.
} 
\vspace{-0.13in}
\centering
\begin{tabular}{lcccccc}
\hline
 & SSIM $\uparrow$ & LPIPS $\downarrow$ & RMSE $\downarrow$ & DISTS $\downarrow$ & PSNR $\uparrow$  \\ \hline
$\losso$                         &0.70 & 0.2563 & 0.1631 & 0.2745 & 21.43 \\
$\losso$ + $\lossp$              &0.75 & 0.1978 & 0.1441 & 0.2005 & 23.26 \\
w/o{} $ \olatsuperresencodernw$  &0.85 & 0.2046 & 0.1465 & 0.1809 &  28.68 \\\hline
$\losso + \lossmrf$              &\textbf{0.87} & \textbf{0.1828} & \textbf{0.1332} & \textbf{0.1689} & \textbf{28.69}  \\\hline
\end{tabular}
\label{tab:ablation-losses-sr_enc}
\end{table}



%
We quantitatively analyze the impact of different loss functions employed in training \OurMethod{} and summarize the results in \cref{tab:ablation-losses-sr_enc}. It is evident that supervision with only the $L1$ loss is insufficient to produce high-quality relighting results. While combining $\lossp$ \cite{zhang2018perceptual} with per-pixel $L1$ loss is a commonly adopted approach, this combination still leads to suboptimal performance, primarily due to this metric missing high-frequency details.  
While $\lossmrf$ alone produces satisfactory results, we find that combining it with $\losso$ further enhances performance and accelerates convergence.

\section{Limitations}\label{sec:limitations}
Despite strong results, \OurMethod{} has several limitations that suggest directions for future work. (i) Although \OurData{} provides full-head coverage, our relighting quality degrades on the back of the head; this stems from the EG3D prior, whose representation does not reliably cover regions outside the front-face region. Integrating a more comprehensive 3D generative prior with \OurData{} could address this limitation and enable full-head relighting. (ii) The scope of this work is limited to facial reflectance (face, eyes, scalp hair). Consequently, headgear and accessories (e.g., helmets, sunglasses) are out of domain, as \OurMethod{} does not synthesize OLATs for these materials. Extending the approach with reflectance priors for a broader set of objects and materials is a promising direction. (iii) Our method inherits EG3D’s difficulty in consistently modeling fine hair fibers: novel-view synthesis can exhibit local inconsistencies in the hair region; small misalignments between OLAT renderings may accumulate into noise or flicker when linearly combined; and the super-resolution stage can introduce strand “popping’’ under head rotation. Addressing these effects will require stronger high-frequency priors and alignment strategies tailored to hair. (iv) Finally, despite conditioning the OLAT decoder $\olatdecnw$ on the viewing direction, view-dependent effects (e.g., on the nose bridge and cheeks) are relatively subdued (see supplementary video). While our OLAT quality (\cref{fig:ablation-olats-wild}) and overall relighting fidelity surpass the baselines, these subtle view-dependent cues contribute weakly to the training objective and are therefore not strongly expressed; improving supervision and objectives for view dependence remains important future work.

\section{Conclusion}
In this paper, we presented \OurMethod{}, a unified framework that addresses the challenge of editing both illumination and viewpoint in portrait images using a single monocular input. Our method draws on the strength of a pre-trained 3D-aware generator, enabling it to learn a rich facial prior.
We used \OurData{}, a new lightstage dataset, in the training of a novel reflectance network, which allows \OurMethod{} to accurately capture facial reflectance through HDR OLAT images.
To enhance the quality of full-head portrait relighting, we use a combination of a reconstruction loss and ID-MRF loss.
Our quantitative and qualitative evaluations show that our method exhibits promising advantages over the existing state-of-the-art approaches, particularly in terms of achieving 3D-consistent editing, simulating accurate light transport effects and controlling novel illumination.
We believe that our work contributes  to the ongoing  research in this field, and we hope it will inspire further exploration and advancements in monocular portrait image editing.

\begin{acks}
This work was supported by the ERC Consolidator Grant 4DReply (770784) and Saarbrücken Research Center for Visual Comput-
ing, Interaction, and AI. 
We thank Oleksandr Sotnychenko for helping us with setting up data capture.
Finally, we thank Shrisha Bharadwaj for discussions, proofreading and innumerable support.
\end{acks}

\bibliographystyle{ACM-Reference-Format}
\bibliography{main}
\balance

\
\pagebreak

\title{Supplementary Material: \OurMethod{} - Single Image 3D Portrait Relighting with Generative Priors}
\section*{Supplementary Material}
We encourage readers to view the supplementary video for a comprehensive display of additional relighting results and OLAT renderings. Upon acceptance of this paper, we will make the \OurData{} dataset, source code and pre-trained weights available.
This supplementary document is organized as follows:
\begin{itemize}
    \item In the \cref{sec:our_dataset} we provide details of our lightstage dataset - \OurData.
    \item In \cref{sec:additional-baselines}, we provide extensive qualitative and quantitative comparisons of \OurMethod~ with state-of-the-art methods in 2D and 3D portrait relighting, demonstrating our robust performance.
    \item In \cref{sec:training_details} we provide training details. 
    \item \Cref{sec:ablations} provide details regarding design ablations, demonstrating how various configurations influence the effectiveness of our method.
\end{itemize}

\section{\OurData~Dataset}
\label{sec:our_dataset}

Capturing accurate, high-fidelity human facial reflectance is essential for advancing research in facial relighting, skin \& hair reflectance modeling, and realistic rendering, particularly for augmented and virtual reality applications. The pioneering work by \citet{debevec2000acquiring} first demonstrated the feasibility and effectiveness of One-Light-At-a-Time (OLAT) setups using specialized hardware, known as a lightstage. Subsequently, the community widely adopted lightstages due to their effectiveness in detailed reflectance capture \cite{sipr_ex, Pandey21, Meka19, mbr_frf, prao2023vorf, zhang2021dynamicolat, saito2024rgca, Bi21, litnerf, sun2020light, mei2024holo, prao2024lite2relight, mallikarjun2021photoapp}. Nevertheless, the complexity, high costs, and demanding calibration procedures of such setups have restricted widespread accessibility. As a consequence, publicly available comprehensive OLAT datasets remain limited, constraining significant progress largely to closed-source initiatives and hindering broader advancements in the field.

Currently, the most extensive publicly available multi-view OLAT dataset is provided by the Codec Avatar Studio \cite{saito2024rgca, martinez2024codec}, consisting of only four relightable subjects captured under dynamic illumination conditions (see Tab. 1 in main paper). This lack of diversity poses significant challenges for data-driven approaches, limiting their ability to generalize effectively to varied facial features and conditions.

To overcome these limitations, we present \OurData, an extensively collected and meticulously calibrated OLAT dataset, distinguished by its scale, diversity, and quality (\cref{fig:dataset_summary}). Unlike prior datasets (refer to Tab. 1 in main paper), \OurData{} uniquely features a large number of multi-view HDR captures at 4K resolution, a substantial variety of subjects, and diverse lighting conditions. These attributes make it particularly suitable for detailed reflectance analysis and the establishment of comprehensive reflectance priors.

In parallel, recent advancements in hair modeling \cite{hair_sklyarova2023haar, hair_Zheng2025GroomLight, hair_zakharov2024gh} have highlighted the critical need for high-quality, multi-view hair datasets, which remain sparse. \OurData{} addresses this specific shortcoming by capturing detailed scalp hair reflectance from multiple viewpoints, encompassing diverse hairstyles, colors, and lengths (\cref{fig:dataset_hair_mv}, \cref{fig:dataset_hair_diversity}, \cref{fig:dataset_summary}). Such comprehensive representation provides valuable priors for improving hair modeling and relighting methods.

\input{figures/tex_figures/dataset_hair}

In the subsequent sections, we detail various aspects of \OurData, including the configuration of the lightstage (\cref{sec:our_dataset_config}), data preprocessing pipeline (\cref{sec:our_dataset_preprocess}), demographic diversity (\cref{subsec:our_dataset_demographics}), and comparative analysis with existing datasets (\cref{sec:our_dataset_compare}). These descriptions aim to clearly illustrate the extensive efforts and specific challenges involved in dataset creation, emphasizing the significant contribution \OurData{} makes to the research community.

\subsection{Lightstage Configuration}
\label{sec:our_dataset_config}
Our capture setup utilizes a 2-meter diameter spherical lightstage equipped with 331 programmable light sources. Each source consists of six LEDs capable of emitting five specific wavelengths: red ($\lambda=630\mathrm{nm}$), green ($\lambda=530\mathrm{nm}$), royal blue ($\lambda=450\mathrm{nm}$), amber ($\lambda=600\mathrm{nm}$), and daylight white ($5650K$).
During the capture sessions, subjects are centrally positioned and uniformly illuminated by white light. The captures are performed synchronously from 40 viewpoints at 4K resolution using RED Komodo cameras operating at 60 FPS.

\begin{figure}
\centering
\newcommand{\mywidth}{1.0}
\includegraphics[]{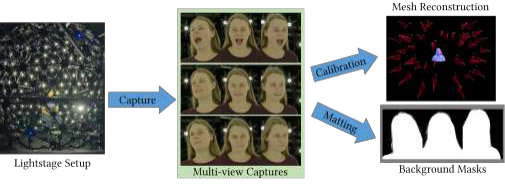}
\caption{\textbf{Data Processing Pipeline.} Each subject is captured under 350 lighting conditions (331 OLAT and 19 fully lit) from 40 viewpoints, across 3 or 4 distinct facial expressions. We then estimate the camera parameters and undistort the images. Finally, background matting techniques are applied to extract background masks.}

\label{fig:dataset_pipeline_supp}
\end{figure}

\subsection{Data Preprocessing}
\label{sec:our_dataset_preprocess}
The captured data undergo rigorous preprocessing to ensure its suitability for model training:

\subsubsection{Optical Flow Alignment}
\label{sec:our_dataset_flow}
\begin{figure}
\centering
\newcommand{\mywidth}{1.0}
\includegraphics[]{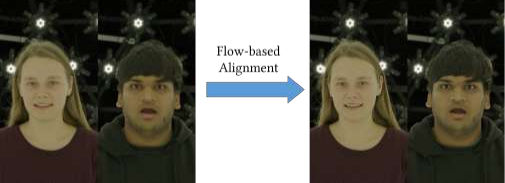}
\caption{\textbf{Impact of Optical Flow-Based Alignment.} We demonstrate that subjects involuntarily tend to introduce minor motion between frames during the capture process, resulting in noticeable blurring when OLAT images are combined (see left column). The results after applying optical flow (see right column) illustrate that correction using RAFT~\cite{Teed2020RAFT} significantly reduces such misalignments.}
\label{fig:dataset_flow_correction}
\end{figure}

Each subject undergoes a sequence of 331 illumination conditions per expression, defined as a \emph{take}, lasting approximately 6 seconds. Minor involuntary subject movements during this interval lead to motion blur when OLAT frames are linearly combined for relighting (\cref{fig:dataset_flow_correction}, left). To address this issue, we intersperse fully lit reference frames every 21 OLAT captures. Optical flow computed using RAFT \cite{Teed2020RAFT} between these reference frames is interpolated linearly to align all OLAT frames, significantly mitigating motion blur, as illustrated in \cref{fig:dataset_pipeline_supp}.

\subsubsection{Background Segmentation}
\label{sec:our_dataset_background}
Accurate foreground-background segmentation is crucial for effective facial reflectance modeling. We generate segmentation masks primarily using BGMv2 \cite{BGMv2}, employing reference background captures without subjects. When BGMv2 produces suboptimal results, we manually select robust alternative masks from RMBGv2 \cite{rmbgv2_BiRefNet}.

\subsubsection{Calibration and 3D Reconstruction}
\label{sec:our_dataset_calibration}
\begin{figure}
\centering
\newcommand{\mywidth}{1.0}
\includegraphics[]{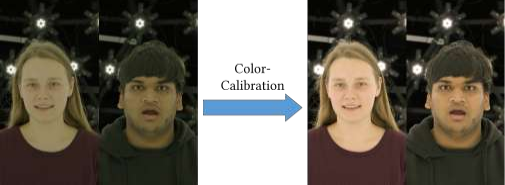}
\caption{\textbf{Color Correction.} \new{We perform color correction on \OurData{}. Shown are representative examples before (left) and after (right) calibration.}}
\label{fig:dataset_color_correction}
\end{figure}

Camera parameters and 3D meshes are reconstructed using Agisoft Metashape\footnote[1]{https://www.agisoft.com/}, with calibration guided by static reference objects and alignment markers placed in the capture volume. We use fully lit frames for reconstruction, as these are devoid of harsh shadows and specular highlights, which are prominent in OLAT images and tend to degrade feature matching quality. Using uniformly illuminated frames improves the robustness of Metashape's feature matching algorithm for both multi-view stereo reconstruction and camera calibration, resulting in an average re-projection error of 0.91 pixels. \new{Further, we will provide a white-balance correction code to correct colors as shown in \cref{fig:dataset_color_correction} along with \OurData{}.}

Despite the superior accuracy achievable through multi-view calibration, we utilize monocular face tracking approaches consistent with monocular portrait relighting methods for our training and evaluations, following EG3D \cite{Chan2022}.

\subsection{Demographics}
\label{subsec:our_dataset_demographics}
We collected a novel multiview lightstage dataset, \OurData, comprising 139 subjects. We made an effort to include participants representing diverse demographics across age, gender, skin color, and hair color. Out of the 139 subjects, 122 participants completed the demographic survey, and their responses are summarized in \cref{fig:dataset_summary}.  

$51.25\%$ of the participants are aged between 24 and 30 years, $17.9\%$ fall within the 31–40 years, $17.9\%$ are between 18 and 23 years, and $9.8\%$ are between 51 and 60 years old. Regarding gender, our dataset includes approximately $59\%$ male, $37\%$ female, and $1.6\%$ non-binary participants, with the remaining subjects opting not to disclose their gender.  

For skin color, we utilize the Fitzpatrick scale\footnote[2]{https://emergetulsa.com/fitzpatrick/}. The dataset comprises $34.1\%$ participants with Type 1 (Light, pale white) skin, $21.1\%$ with Type 2 (White, fair) skin, $20.3\%$ with Type 3 (Medium, white to olive) skin, $14.6\%$ with Type 4 (Olive, moderate brown) skin, $4.9\%$ with Type 5 (Brown, dark brown), and $4.9\%$ with Type 6 (Black, very dark brown to black) skin.  

Hair color is reported using a  hair color scale\footnote[3]{https://satinhaircolor.com/education/}, where higher values correspond to lighter hair colors, and lower values represent darker hair tones. 

\input{figures/tex_figures/dataset_summary}

\subsection{Comparison with Existing Datasets}
\label{sec:our_dataset_compare}

In this section, we highlight the key features that make \OurData~ a unique contribution compared to existing publicly available light stage datasets. \OurData~ offers the following advantages:

\begin{itemize}
  \item \textbf{Illumination Coverage:} 331 lighting conditions, each corresponding to a unique incident direction, enabling detailed relighting effects such as specular highlights, subsurface scattering, self-shadowing, and hard shadows.
  
  \item \textbf{Scale and Diversity:} Our corpus comprises 139 subjects, providing a diverse set of facial geometries and appearances. This results in enhanced generalizability and constitutes the \textbf{largest database} of relightable heads available to date.
  
  \item \textbf{Multi-view Resolution:} Consistent 4K resolution across 40 viewpoints, capturing high-frequency details in facial features, hair, and eye regions.
\end{itemize}

The most comparable dataset to ours is the dynamic face OLAT dataset from Codec Avatar Studio~\cite{saito2024rgca, martinez2024codec}, which includes only 4 subjects captured under grouped OLAT lighting conditions. This dataset suffers from two primary limitations: (1) the small number of subjects limits its ability to support robust face reflectance prior modeling, and (2) grouped OLAT captures are less suitable for static image-based relighting due to reduced directional specificity.

Another relevant dataset is Ultrastage~\cite{zhou2023relightable}, which captures full-body subjects under gradient illumination. While gradient lighting is effective for estimating photometric normals, it is suboptimal for high-quality relighting. Gradient-based relighting approaches, such as Ultrastage, rely on simplified BRDF assumptions and fail to capture complex light-skin interactions. In contrast, our OLAT-based acquisition captures the complete reflectance field, making it highly suitable for physically accurate, image-based relighting.

Additional publicly available datasets, including ICT-3DFE~\cite{stratou2011ICTFaceDB} and the Dynamic OLAT dataset~\cite{zhang2021dynamicolat}, capture subjects from only one or two viewpoints, thereby limiting their utility for 3D face modeling. Furthermore, these datasets contain few subjects and are limited in resolution.

Finally, although not publicly available, the OLAT dataset by Weyrich et al.~\cite{Weyrich2006Analysis} has long served as a standard benchmark for evaluating face reflectance models\cite{mbr_frf, Weyrich2006Analysis, mallikarjun2021photoapp, prao2022vorf, prao2023vorf, prao2024lite2relight, sun2021nelf}. Hence, we train and compare 3DPR and prior methods on this dataset to ensure standardized and fair evaluation. However, it presents several key limitations: (1) it is a closed-source dataset, limiting reproducibility and broader research adoption; (2) all subjects exhibit a single, neutral expression with closed eyes, making it unsuitable for modeling eye reflectance, expressions, and mouth interior; and (3) it captures only the frontal face hemisphere from 14 views, with no coverage of hair or the back of the head. \OurData~ addresses each of these shortcomings, offering public access, diverse expressions with open-eye captures, and full-head coverage including hair, thus providing a more comprehensive resource for developing and evaluating face reflectance models.

In summary, \OurData~ addresses critical limitations of existing datasets by providing a large-scale, high-resolution, and multi-view dataset for full-head reflectance modeling, making it a valuable resource for the community.

\section{Additional Baselines}\label{sec:additional-baselines}
In this section, we first provide qualitative comparisons with Holo-Relighting~\cite{mei2024holo} (HR). Next, we evaluate our approach against a 2D portrait relighting method, Total Relighting~\cite{Pandey21}. Additionally, we present further qualitative evaluations for Light2Relight, NeRFFaceLighting, VoRF, and NeLF.

\subsection{Holo-Relighting}\label{subsec:holo}
We provide a comparison with Holo-Relighting~\cite{mei2024holo}. Their method also relies on an EG3D prior to reconstruct the facial volume for view synthesis and is trained on lightstage data for relighting. However, unlike our approach, their pipeline directly inputs an environment map and produces the relit result without generating OLATs, leading to physically implausible outputs.
As shown in \cref{fig:baseline_holo_relit}, their technique often produces oversaturated results and lacks finer details, such as facial hair. Furthermore, as evidenced in \cref{fig:baseline_holo_olat}, Holo-Relighting struggles to accurately relight under point light conditions, despite being trained on a lightstage dataset. For example, it fails to reproduce complex skin-light interactions, such as subsurface scattering (see the ear regions in columns 1 and 2). Additionally, under a point light source positioned below the face, Holo-Relighting fails to generate coherent relit results.
In contrast, \OurMethod{} effectively handles subsurface scattering effects and achieves accurate relighting, as evidenced by its close match to the reference image.

Since neither the code nor implementation details of HR are publicly available, retraining their method with our data for quantitative evaluation was not feasible. Instead, we shared our evaluation data with the authors. Upon discussion, we found that quantitative comparisons would be challenging due to differences in the radiance scaling factors for the HDR environment maps and variations in lightstage capture setups. Therefore, we provide only qualitative comparisons by evaluating Holo-Relighting against \OurMethod{} trained on \OurData{}.

\input{figures/tex_figures/baseline_holo_relit}
\input{figures/tex_figures/baseline_holo_olat}

\subsection{Total Relighting}
\begin{table}[t]
\centering
\caption{\textbf{Quantitative comparison with Total Relighting\cite{Pandey21}:} Metrics are evaluated for relighting performance across 10 unseen lightstage subjects \cite{Weyrich2006Analysis}  under 27 different illumination conditions.}
\scalebox{0.8}{
\begin{tabular}{lcccccc}
\hline
 & LPIPS $\downarrow$ & RMSE $\downarrow$ & DISTS $\downarrow$ & SSIM $\uparrow$ & PSNR $\uparrow$ \\ \hline
Total Relighting  &0.2991 &0.2203 & 0.2228 & 0.8612 & \textbf{29.25} \\\hline
Ours ($\OldModel$) &\textbf{0.2599} &\textbf{0.1660} &\textbf{0.1933} & \textbf{0.8772} & {28.73} \\\hline
\end{tabular}
}
\label{tab:baseline_TR}
\end{table}
Total Relighting \cite{Pandey21} (TR) explicitly decomposes face reflectance by employing dedicated networks to separately extract surface normals, albedo, and specular maps. This method utilizes the Phong reflectance function to facilitate the decomposition process. 
We conduct both qualitative and quantitative comparisons of \OurMethod~ with Total Relighting (TR) to assess the effectiveness of our proposed relighting approach. In our quantitative analysis, we evaluate the performance using metrics such as LPIPS \cite{zhang2018perceptual}, RMSE, DISTS \cite{dists_metric}, SSIM, and PSNR across 10 unseen lightstage subjects under 27 different illumination conditions. For TR, due to the lack of publicly available code or models, evaluations were performed by original authors using their pre-trained models using our inputs.

The results, detailed in \cref{tab:baseline_TR}, demonstrate that \OurMethod~ surpasses the TR baseline in all metrics except PSNR, which achieves a score closely competitive with TR. This underscores our method's high-quality performance, aligning closely with the state-of-the-art while potentially offering additional advantages \ie relighting under novel views.
Building on our quantitative assessments, qualitative evaluations from \cref{fig:baseline_TR_LS}, and \cref{fig:baseline_TR_wild_faceolat} further illustrate the advantages of \OurMethod~ over TR, in handling both lightstage and in-the-wild subjects. Our method is successful in capturing accurate shadows and specular highlights, in contrast to TR, which often exhibits issues with inaccurate lighting and produces cloudy artifacts on facial regions.
These observations highlight the critical role of employing a rich face prior and utilizing implicit face reflectance modeling through OLATs. These strategies enable \OurMethod~ to achieve precise relighting across diverse illumination conditions.

\input{figures/tex_figures/baseline_TR_LS}
\input{figures/tex_figures/baseline_TR_wild_faceolat}

\subsection{IC-Light}
\new{We compare against the recent diffusion-based relighting method IC-Light~\cite{zhang2025_ic_light}. Diffusion models provide strong priors for generalization; however, we observe that IC-Light, while producing plausible results, struggles to reproduce physically accurate lighting effects such as specular highlights, subsurface scattering, and cast shadows. These limitations are more pronounced under sparse, highly directional illumination (see input~2 in \cref{fig:ic_light-wild_novel_view}). We also evaluate cross-view relighting consistency by first generating novel viewpoints with EG3D and then relighting them using IC-Light; the results reveal inconsistent lighting across viewpoints (\cref{fig:ic_light-wild_novel_view}). Additionally, quantitative results on FaceOLAT (\cref{tab:baseline_ICL}) show that \OurMethod{} outperforms IC-Light, including stronger identity preservation during relighting. We attribute the superior performance of \OurMethod{} to its 3D-consistent generative prior coupled with explicit OLAT-based reflectance modeling notably absent in IC-Light.}

\begin{table}[h]
\centering
\caption{\textbf{Quantitative comparison with IC-Light\cite{zhang2025_ic_light}:} \new{Metrics are evaluated for relighting performance on \OurData{} under 10 different illumination conditions.}}
\scalebox{0.8}{
\begin{tabular}{lcccccc}
\hline
 & SSIM $\uparrow$ & LPIPS $\downarrow$ & RMSE $\downarrow$ & DISTS $\downarrow$ & PSNR $\uparrow$ & ID $\uparrow$ \\ \hline
IC-Light (input-view)  &0.70 &0.309 & 0.375 & 0.284 & 20.36 & 0.913 \\\hline
IC-Light (novel-view) &0.67 &0.313 & 0.470 & 0.309 & 20.24 & 0.874 \\\hline
Ours &\textbf{0.83} &\textbf{0.199} &\textbf{0.180} & \textbf{0.175} & \textbf{21.02} & \textbf{0.943} \\\hline
\end{tabular}
}
\label{tab:baseline_ICL}
\end{table}
\input{figures/tex_figures/baseline_IC_light_multi_view}

\subsection{NeRFFaceLighting and Lite2Relight}
\label{subsec:nfl}
In this section, we provide additional qualitative comparisons for: NeRFFaceLighting (NFL)~\cite{nerffacelighting} and Lite2Relight (L2R)~\cite{prao2024l2r} in \cref{fig:nfl_l2r-wild_novel_view} and \cref{fig:baseline_nfl_l2r-wild_same_view}.
We observe that both L2R and NFL begin to degrade under sparse illumination conditions. NFL struggles significantly with identity preservation and lighting realism. Its spherical harmonics (SH)-based lighting representation captures only low-frequency cues, and it fails to model high-frequency effects such as sharp shadows or specular highlights. In addition, we observe that lighting tends to be baked into the albedo during its inaccurate inversion process, further limiting generalization under colored lighting conditions.

L2R performs reasonably well under dense natural lighting but fails under sparse or highly directional environments, likely due to its latent-space probing strategy, which is biased by the EG3D training distribution. This leads to artifacts such as inconsistent illumination, lack of shadows, and distorted identity features in extreme cases.  
To verify this observation, we contacted the authors of Lite2Relight, who confirmed that this is a known limitation of their method, as documented in the original paper.

In contrast, our method explicitly models  face reflectance using OLATs and linearly combines these basis images with arbitrary environment maps. This allows for faithful and consistent relighting, even under sparse or colored light setups. The figure clearly illustrates our method’s ability to maintain photorealism, preserve facial structure, and accurately reproduce lighting effects across a wide range of conditions. These results highlight the importance and effectiveness of our OLAT-based formulation for generalizable relighting.

\input{figures/tex_figures/baseline_new_SOTA_input_view}
\input{figures/tex_figures/baseline_new_SOTA_multi_view}

\subsubsection{NeLF}
\input{figures/tex_figures/baseline_nelf}
NeLF \cite{sun2021nelf} proposes a pixelNeRF-based \cite{yu2020pixelnerf} method that leverages image features to infer the underlying 3D geometry and light transport.
We present qualitative results for NeLF in \cref{fig:baseline-NeLF}. We can see that despite utilizing three input views, NeLF encounters difficulties in achieving coherent face reconstruction. 
This issue arises from its reliance on the aggregation of local image features to infer geometry, which results in poor facial geometry reconstruction. Consequently, NeLF scores poorly as well as reported in the Tab.2 of the main paper.

\subsection{VoRF}
\input{figures/tex_figures/baseline_VORF}
\begin{table}[t]
\caption{\textbf{Quantitative Results: OLAT Evaluation} We report the quality of synthesized OLATs for simultaneous view-synthesis and relighting.}
\label{tab:ablation-olats}
\centering
\begin{tabular}{lcccc}
\hline
 & SSIM $\uparrow$ &  LPIPS $\downarrow$ & PSNR $\uparrow$ \\ \hline
VoRF  & 0.71 & 0.3148 & 20.43 \\\hline
Ours  &\textbf{0.88} &\textbf{0.1753} &\textbf{28.70} \\\hline
\end{tabular}
\end{table}
For qualitative analysis, we compare $\OldModel$ with VoRF, both trained on \OldData{}, and present the corresponding quantitative results in \cref{tab:ablation-olats}, where \OurMethod{} significantly outperforms VoRF across all metrics.
Since both methods rely on explicit OLAT representations to model facial reflectance, it is crucial to assess the accuracy of the synthesized OLATs. As shown in \cref{fig:baseline-VORF}, \OurMethod{} generalizes effectively to both our evaluation dataset (2nd row) and to \enquote{in-the-wild} subjects (3rd and 4th rows), producing higher-fidelity relighting results that capture complex lighting effects more reliably than VoRF.
Note that for evaluating VoRF against 3DPR, we use $\OldModel$. Since the lightstage dataset of \citet{Weyrich2006Analysis} contains subjects with closed eyes, $\OldModel$ fails to accurately model the reflectance of the eye region. Consequently, we mask the eye region in \cref{fig:baseline-VORF} and \cref{fig:ablation-sr_enc}.

\subsection{LumiGAN }\label{subsec:lumingan}
LumiGAN~\cite{deng2023lumigan} decomposes a given face into geometry, albedo, specular, and visibility maps. Next, it combines the decomposed rendering with spherical harmonic projections of desired environment maps to perform relighting. 
The method relies entirely on adversarial training which produces plausible results, but, due to the lack of relit ground truth supervision during training, LumiGAN lacks physically accurate lighting. For instance, by observing \textbf{Figure S3} \cite{deng2023lumigan} we can see that the direction of illumination and specularities (observe the highlights on forehead and nose region) are inconsistent across subjects for the same environment map. Moreover, the specular highlights appear in a few subjects, but are missing in others. We hypothesize that these occurrences are due to inaccurate decomposition of face reflectance components. 

In contrast, \OurMethod~is based on \emph{volumetric reflectance fields}. Our face reflectance modeling is built using OLATs and these OLATs are captured in a lightstage~\cite{Weyrich2006Analysis} setup that can represent physically accurate lighting. Thus, by supervising our method with OLAT dataset as ground truth, we learn physically accurate lighting.   Since neither the code, nor the pretrained weights of LumiGAN are publicly available, we have compared our method with a state-of-the-art GAN-based approach that is similar to LumiGAN in formulation: NeRFFaceLighting \cite{nerffacelighting} (see main paper for the results).

\subsection{VoLux-GAN}
VoLux-GAN~\cite{tan2022voluxgan} is a 3D-aware generator capable of relighting arbitrary samples, but cannot reconstruct and relight a specific given portrait image. Conversations with the VoLux-GAN authors confirmed that the rich StyleGAN latent space, which facilitates the embedding of test subjects, is not preserved in their approach due to the disentangled representation. This is further reinforced by the fact that they do not show any inversion or reconstruction results, but only generated identities sampled from the latent space.
\OurMethod, however, is not just another 3D-GAN approach, but specifically designed for portrait relighting, capable of reconstructing and relighting given portrait images, which VoluxGAN cannot do. 

\section{Training} \label{sec:training_details}
\new{
We train 3DPR with 100 HDRIs (60 indoor, 40 outdoor) per subject, covering a wide range of color temperatures and lighting conditions, which makes it robust to in-the-world lighting during inference as demonstrated by our superior performance in both qualitative and quantiative evaluations.}

\OurMethod{} is implemented using PyTorch and integrates a pre-trained 3D generator \cite{Chan2022} alongside a geometry-aware inversion framework \cite{goae}. 
We train our method on two different lightstage datasets: \OurData{} and $\OldData$~\cite{Weyrich2006Analysis}. The trained models are denoted as $\OurModel$ and $\OldModel$, respectively, in our results.
During the first 60k training iterations, we only train $\olatencodernw$ and $\olatdecnw$, supervised by low-resolution OLAT images $\olatrgb$. 
This "warm-up" phase is crucial for stabilizing the training process.
Following this,  $\olatsuperresencodernw{}$ is initialized with weights from the original EG3D super-resolution network. 
We then proceed to training all modules jointly, including  $\olatsuperresencodernw$, for an additional 150k iterations.
Training uses a batch size of 8 on 4 $\times$ NVIDIA H100 GPUs over a span of 2 days. 
We employ the Adam optimizer \cite{KingmaB14} with a learning rate of $0.00015$. 
The training dataset consists of 130 subjects, each re-illuminated under 20 randomly sampled natural illumination conditions from three different viewpoints.

\section{Design Ablation Study}~\label{sec:ablations}
In this section, we ablate various design choices of \OurMethod{} qualitatively and quantitatively. We conduct all the ablations using $\OldModel$ that was trained on $\OldData$ by \citet{Weyrich2006Analysis}.

\subsection{Significance of SR Encoder:}
\input{figures/tex_figures/ablation_sr_enc}
We summarize qualitative analysis in the \Cref{fig:ablation-sr_enc} to highlight the importance of $\olatsuperresencodernw$.

\subsection{Significance of $\lossmrf$:}
We present the qualitative results \cref{fig:ablation-losses} clearly shows that including $\lossmrf$ in training leads to best results.
\input{figures/tex_figures/ablation_losses}

\subsection{Effect of Training Subjects Number}
\begin{table}[t]
\centering
\caption{\textbf{Quantitative Results: Number of Training Subjects}. 
The performance metrics are evaluated for relighting performance across 
10 unseen Lightstage subjects \cite{Weyrich2006Analysis}.}
\begin{tabular}{lccc}
\hline
       & SSIM $\uparrow$ & PSNR $\uparrow$ & LD $\downarrow$ \\
\hline
$n=5$  & 0.85 & 28.66 & 11.18 \\
$n=25$ & 0.86 & 28.62 & 10.71 \\
$n=100$& 0.86 & 28.61 & 10.56 \\
\hline
$n=250$& \textbf{0.87} & \textbf{28.69} & \textbf{10.3} \\
\hline
\end{tabular}
\label{tab:ablation-n_subjects}
\end{table}

\input{figures/tex_figures/ablation_n_subjects}
We evaluate the task of simultaneous view synthesis and relighting using multiple models trained with $250$, $100$, $25$ and as few as $5$ subjects. We observe that with as few as 5 subjects we achieve reasonable relighting shown in \Cref{fig:ablation-n_subjects}. Although we achieve the best performance with $250$ subjects. From the quantitative results in  \Cref{tab:ablation-n_subjects} we observe a similar performance with $100$ and $25$ training subjects. Further, we achieve relighting without involving any tedious optimization process compared to VoRF \cite{prao2023vorf} or NFL\cite{nerffacelighting}. This demonstrates the significance of the EG3D prior that helps in generalization towards unseen subjects including in-the-wild captures (see Fig. 3 in the main paper). In, \cref{tab:ablation-n_subjects} we report an average facial landmarks deviation (LD) metric to evaluate facial geometry consistency by measuring the average deviation of 68 facial key points \cite{bulat2017far}.

\subsection{Dimensionality of \olatencoder}
\begin{table}[t]
\centering
\caption{\textbf{Quantitative Results: Ablation Study on $\olatfeaturemap$ dimension} .
The performance metrics are evaluated for relighting performance across 
10 unseen Lightstage subjects \cite{Weyrich2006Analysis}.
}
\begin{tabular}{lcccc}
\hline
 & SSIM $\uparrow$ & PSNR $\uparrow$ & LD $\downarrow$\\ \hline
$D=9$  &0.84 &22.62 & 10.89\\
$D=48$ &0.86 &28.74 & 11.57 \\\hline
$D=96$ &\textbf{0.87} &\textbf{28.69} &\textbf{10.30}\\\hline
\end{tabular}
\label{tab:ablation-n_dim}
\end{table}
\input{figures/tex_figures/ablation_n_dim}
The pre-trained EG3D triplane representation, $\egtdfeaturemap$, consists of 96 dimensions. In this ablation study, we investigate the impact of varying the dimension size of $\olatfeaturemap$ on relighting quality. We start with a depth of $\olatfeaturemap$ at 9 dimensions. At this level, without sufficient $\olathf$ to encode high-frequency details, the relighting results appear blurry, as illustrated in \Cref{fig:ablation-n_dim}.
As the depth of $\olatfeaturemap$ increases, we observe enhancements in the quality of relighting and facial features. The best results are achieved when $\olatfeaturemap$ is extended to 96 dimensions, as detailed in \Cref{tab:ablation-n_dim}. This configuration allows for effective encoding of complex lighting effects, including subsurface scattering (visible in the first row's third and fourth columns) and specular highlights on the nose (noted in the second row's fourth column).

\bibliographystyle{ACM-Reference-Format}
\bibliography{main}


\begin{thebibliography}{75}


\ifx \showCODEN    \undefined \def \showCODEN     #1{\unskip}     \fi
\ifx \showDOI      \undefined \def \showDOI       #1{#1}\fi
\ifx \showISBNx    \undefined \def \showISBNx     #1{\unskip}     \fi
\ifx \showISBNxiii \undefined \def \showISBNxiii  #1{\unskip}     \fi
\ifx \showISSN     \undefined \def \showISSN      #1{\unskip}     \fi
\ifx \showLCCN     \undefined \def \showLCCN      #1{\unskip}     \fi
\ifx \shownote     \undefined \def \shownote      #1{#1}          \fi
\ifx \showarticletitle \undefined \def \showarticletitle #1{#1}   \fi
\ifx \showURL      \undefined \def \showURL       {\relax}        \fi
\providecommand\bibfield[2]{#2}
\providecommand\bibinfo[2]{#2}
\providecommand\natexlab[1]{#1}
\providecommand\showeprint[2][]{arXiv:#2}

\bibitem[{B R} et~al\mbox{.}(2021a)]%
        {mallikarjun2021photoapp}
\bibfield{author}{\bibinfo{person}{Mallikarjun {B R}}, \bibinfo{person}{Ayush
  Tewari}, \bibinfo{person}{Abdallah Dib}, \bibinfo{person}{Tim Weyrich},
  \bibinfo{person}{Bernd Bickel}, \bibinfo{person}{Hans~Peter Seidel},
  \bibinfo{person}{Hanspeter Pfister}, \bibinfo{person}{Wojciech Matusik},
  \bibinfo{person}{Louis Chevallier}, \bibinfo{person}{Mohamed~A Elgharib},
  {and} \bibinfo{person}{Christian Theobalt}.}
  \bibinfo{year}{2021}\natexlab{a}.
\newblock \showarticletitle{PhotoApp: Photorealistic appearance editing of head
  portraits}.
\newblock \bibinfo{journal}{\emph{ACM Transactions on Graphics}}
  \bibinfo{volume}{40}, \bibinfo{number}{4} (\bibinfo{year}{2021}).
\newblock


\bibitem[{B R} et~al\mbox{.}(2021b)]%
        {mbr_frf}
\bibfield{author}{\bibinfo{person}{Mallikarjun {B R}}, \bibinfo{person}{Ayush
  Tewari}, \bibinfo{person}{Tae-Hyun Oh}, \bibinfo{person}{Tim Weyrich},
  \bibinfo{person}{Bernd Bickel}, \bibinfo{person}{Hans-Peter Seidel},
  \bibinfo{person}{Hanspeter Pfister}, \bibinfo{person}{Wojciech Matusik},
  \bibinfo{person}{Mohamed Elgharib}, {and} \bibinfo{person}{Christian
  Theobalt}.} \bibinfo{year}{2021}\natexlab{b}.
\newblock \showarticletitle{Monocular Reconstruction of Neural Face Reflectance
  Fields}. In \bibinfo{booktitle}{\emph{The IEEE Conference on Computer Vision
  and Pattern Recognition (CVPR)}}.
\newblock


\bibitem[Bi et~al\mbox{.}(2021)]%
        {Bi21}
\bibfield{author}{\bibinfo{person}{Sai Bi}, \bibinfo{person}{Stephen Lombardi},
  \bibinfo{person}{Shunsuke Saito}, \bibinfo{person}{Tomas Simon},
  \bibinfo{person}{Shih-En Wei}, \bibinfo{person}{Kevyn Mcphail},
  \bibinfo{person}{Ravi Ramamoorthi}, \bibinfo{person}{Yaser Sheikh}, {and}
  \bibinfo{person}{Jason Saragih}.} \bibinfo{year}{2021}\natexlab{}.
\newblock \showarticletitle{Deep Relightable Appearance Models for Animatable
  Faces}.
\newblock \bibinfo{journal}{\emph{ACM Transactions on Graphics}}, Article
  \bibinfo{articleno}{89} (\bibinfo{year}{2021}), \bibinfo{numpages}{15}~pages.
\newblock


\bibitem[Boss et~al\mbox{.}(2021)]%
        {Boss20NeRD}
\bibfield{author}{\bibinfo{person}{Mark Boss}, \bibinfo{person}{Raphael Braun},
  \bibinfo{person}{Varun Jampani}, \bibinfo{person}{Jonathan~T. Barron},
  \bibinfo{person}{Ce Liu}, {and} \bibinfo{person}{Hendrik~P.A. Lensch}.}
  \bibinfo{year}{2021}\natexlab{}.
\newblock \showarticletitle{NeRD: Neural Reflectance Decomposition from Image
  Collections}. In \bibinfo{booktitle}{\emph{The IEEE International Conference
  on Computer Vision (ICCV)}}.
\newblock


\bibitem[Buehler et~al\mbox{.}(2021)]%
        {buehler2021varitex}
\bibfield{author}{\bibinfo{person}{Marcel~C. Buehler},
  \bibinfo{person}{Abhimitra Meka}, \bibinfo{person}{Gengyan Li},
  \bibinfo{person}{Thabo Beeler}, {and} \bibinfo{person}{Otmar Hilliges}.}
  \bibinfo{year}{2021}\natexlab{}.
\newblock \showarticletitle{VariTex: Variational Neural Face Textures}. In
  \bibinfo{booktitle}{\emph{Proceedings of the IEEE/CVF International
  Conference on Computer Vision}}.
\newblock


\bibitem[Bulat and Tzimiropoulos(2017)]%
        {bulat2017far}
\bibfield{author}{\bibinfo{person}{Adrian Bulat} {and}
  \bibinfo{person}{Georgios Tzimiropoulos}.} \bibinfo{year}{2017}\natexlab{}.
\newblock \showarticletitle{How far are we from solving the 2D \& 3D Face
  Alignment problem? (and a dataset of 230,000 3D facial landmarks)}. In
  \bibinfo{booktitle}{\emph{International Conference on Computer Vision}}.
\newblock


\bibitem[Chan et~al\mbox{.}(2022)]%
        {Chan2022}
\bibfield{author}{\bibinfo{person}{Eric~R. Chan}, \bibinfo{person}{Connor~Z.
  Lin}, \bibinfo{person}{Matthew~A. Chan}, \bibinfo{person}{Koki Nagano},
  \bibinfo{person}{Boxiao Pan}, \bibinfo{person}{Shalini~De Mello},
  \bibinfo{person}{Orazio Gallo}, \bibinfo{person}{Leonidas Guibas},
  \bibinfo{person}{Jonathan Tremblay}, \bibinfo{person}{Sameh Khamis},
  \bibinfo{person}{Tero Karras}, {and} \bibinfo{person}{Gordon Wetzstein}.}
  \bibinfo{year}{2022}\natexlab{}.
\newblock \showarticletitle{Efficient Geometry-aware {3D} Generative
  Adversarial Networks}. In \bibinfo{booktitle}{\emph{CVPR}}.
\newblock


\bibitem[Chandran et~al\mbox{.}(2022)]%
        {Chandran_2022_WACV}
\bibfield{author}{\bibinfo{person}{Sreenithy Chandran},
  \bibinfo{person}{Yannick Hold-Geoffroy}, \bibinfo{person}{Kalyan Sunkavalli},
  \bibinfo{person}{Zhixin Shu}, {and} \bibinfo{person}{Suren Jayasuriya}.}
  \bibinfo{year}{2022}\natexlab{}.
\newblock \showarticletitle{Temporally Consistent Relighting for Portrait
  Videos}. In \bibinfo{booktitle}{\emph{The IEEE Winter Conference on
  Applications of Computer Vision (WACV) Workshops}}.
  \bibinfo{pages}{719--728}.
\newblock


\bibitem[Debevec et~al\mbox{.}(2000)]%
        {debevec2000acquiring}
\bibfield{author}{\bibinfo{person}{Paul Debevec}, \bibinfo{person}{Tim
  Hawkins}, \bibinfo{person}{Chris Tchou}, \bibinfo{person}{Haarm-Pieter
  Duiker}, \bibinfo{person}{Westley Sarokin}, {and} \bibinfo{person}{Mark
  Sagar}.} \bibinfo{year}{2000}\natexlab{}.
\newblock \showarticletitle{Acquiring the reflectance field of a human face}.
  In \bibinfo{booktitle}{\emph{Annual conference on Computer graphics and
  interactive techniques}}.
\newblock


\bibitem[Deng et~al\mbox{.}(2023)]%
        {deng2023lumigan}
\bibfield{author}{\bibinfo{person}{Boyang Deng}, \bibinfo{person}{Yifan Wang},
  {and} \bibinfo{person}{Gordon Wetzstein}.} \bibinfo{year}{2023}\natexlab{}.
\newblock \showarticletitle{LumiGAN: Unconditional Generation of Relightable 3D
  Human Faces}. In \bibinfo{booktitle}{\emph{arXiv}}.
\newblock


\bibitem[Deng et~al\mbox{.}(2022)]%
        {gram}
\bibfield{author}{\bibinfo{person}{Yu Deng}, \bibinfo{person}{Jiaolong Yang},
  \bibinfo{person}{Jianfeng Xiang}, {and} \bibinfo{person}{Xin Tong}.}
  \bibinfo{year}{2022}\natexlab{}.
\newblock \showarticletitle{GRAM: Generative Radiance Manifolds for 3D-Aware
  Image Generation}. In \bibinfo{booktitle}{\emph{IEEE/CVF Conference on
  Computer Vision and Pattern Recognition}}.
\newblock


\bibitem[Ding et~al\mbox{.}(2020)]%
        {dists_metric}
\bibfield{author}{\bibinfo{person}{Keyan Ding}, \bibinfo{person}{Kede Ma},
  \bibinfo{person}{Shiqi Wang}, {and} \bibinfo{person}{Eero~P. Simoncelli}.}
  \bibinfo{year}{2020}\natexlab{}.
\newblock \showarticletitle{Image Quality Assessment: Unifying Structure and
  Texture Similarity}.
\newblock \bibinfo{journal}{\emph{CoRR}}  \bibinfo{volume}{abs/2004.07728}
  (\bibinfo{year}{2020}).
\newblock
\urldef\tempurl%
\url{https://arxiv.org/abs/2004.07728}
\showURL{%
\tempurl}


\bibitem[Goodfellow et~al\mbox{.}(2014)]%
        {goodfellow2014generative}
\bibfield{author}{\bibinfo{person}{Ian Goodfellow}, \bibinfo{person}{Jean
  Pouget-Abadie}, \bibinfo{person}{Mehdi Mirza}, \bibinfo{person}{Bing Xu},
  \bibinfo{person}{David Warde-Farley}, \bibinfo{person}{Sherjil Ozair},
  \bibinfo{person}{Aaron Courville}, {and} \bibinfo{person}{Yoshua Bengio}.}
  \bibinfo{year}{2014}\natexlab{}.
\newblock \showarticletitle{Generative adversarial nets}. In
  \bibinfo{booktitle}{\emph{Advances in neural information processing
  systems}}. \bibinfo{pages}{2672--2680}.
\newblock


\bibitem[Gu et~al\mbox{.}(2022)]%
        {gu2021stylenerf}
\bibfield{author}{\bibinfo{person}{Jiatao Gu}, \bibinfo{person}{Lingjie Liu},
  \bibinfo{person}{Peng Wang}, {and} \bibinfo{person}{Christian Theobalt}.}
  \bibinfo{year}{2022}\natexlab{}.
\newblock \showarticletitle{StyleNeRF: A Style-based 3D Aware Generator for
  High-resolution Image Synthesis}. In \bibinfo{booktitle}{\emph{International
  Conference on Learning Representations}}.
\newblock


\bibitem[Haotian et~al\mbox{.}(2024)]%
        {yang2024vrmm}
\bibfield{author}{\bibinfo{person}{Yang Haotian}, \bibinfo{person}{Zheng
  Mingwu}, \bibinfo{person}{Ma ChongYang}, \bibinfo{person}{Lai Yu-Kun},
  \bibinfo{person}{Wan Pengfei}, {and} \bibinfo{person}{Huang Haibin}.}
  \bibinfo{year}{2024}\natexlab{}.
\newblock \showarticletitle{VRMM: A Volumetric Relightable Morphable Head
  Model}. In \bibinfo{booktitle}{\emph{SIGGRAPH 2024 Conference Proceedings}}.
\newblock


\bibitem[He et~al\mbox{.}(2016)]%
        {he2016residual}
\bibfield{author}{\bibinfo{person}{Kaiming He}, \bibinfo{person}{Xiangyu
  Zhang}, \bibinfo{person}{Shaoqing Ren}, {and} \bibinfo{person}{Jian Sun}.}
  \bibinfo{year}{2016}\natexlab{}.
\newblock \showarticletitle{{Deep Residual Learning for Image Recognition}}. In
  \bibinfo{booktitle}{\emph{Proceedings of 2016 IEEE Conference on Computer
  Vision and Pattern Recognition}} (Las Vegas, NV, USA)
  \emph{(\bibinfo{series}{CVPR '16})}. \bibinfo{publisher}{IEEE},
  \bibinfo{pages}{770--778}.
\newblock
\showISSN{1063-6919}
\urldef\tempurl%
\url{https://doi.org/10.1109/CVPR.2016.90}
\showDOI{\tempurl}


\bibitem[He et~al\mbox{.}(2024)]%
        {diffrelight_he}
\bibfield{author}{\bibinfo{person}{Mingming He}, \bibinfo{person}{Pascal
  Clausen}, \bibinfo{person}{Ahmet~Levent Ta\c{s}el}, \bibinfo{person}{Li Ma},
  \bibinfo{person}{Oliver Pilarski}, \bibinfo{person}{Wenqi Xian},
  \bibinfo{person}{Laszlo Rikker}, \bibinfo{person}{Xueming Yu},
  \bibinfo{person}{Ryan Burgert}, \bibinfo{person}{Ning Yu}, {and}
  \bibinfo{person}{Paul Debevec}.} \bibinfo{year}{2024}\natexlab{}.
\newblock \showarticletitle{DifFRelight: Diffusion-Based Facial Performance
  Relighting}. In \bibinfo{booktitle}{\emph{SIGGRAPH Asia 2024 Conference
  Papers}} \emph{(\bibinfo{series}{SA '24})}. \bibinfo{publisher}{Association
  for Computing Machinery}, \bibinfo{address}{New York, NY, USA}, Article
  \bibinfo{articleno}{11}, \bibinfo{numpages}{12}~pages.
\newblock
\showISBNx{9798400711312}
\urldef\tempurl%
\url{https://doi.org/10.1145/3680528.3687644}
\showDOI{\tempurl}


\bibitem[Hong et~al\mbox{.}(2022)]%
        {headnerf}
\bibfield{author}{\bibinfo{person}{Yang Hong}, \bibinfo{person}{Bo Peng},
  \bibinfo{person}{Haiyao Xiao}, \bibinfo{person}{Ligang Liu}, {and}
  \bibinfo{person}{Juyong Zhang}.} \bibinfo{year}{2022}\natexlab{}.
\newblock \showarticletitle{HeadNeRF: A Real-time NeRF-based Parametric Head
  Model}. In \bibinfo{booktitle}{\emph{{IEEE/CVF} Conference on Computer Vision
  and Pattern Recognition (CVPR)}}.
\newblock


\bibitem[Jiang et~al\mbox{.}(2023)]%
        {nerffacelighting}
\bibfield{author}{\bibinfo{person}{Kaiwen Jiang}, \bibinfo{person}{Shu-Yu
  Chen}, \bibinfo{person}{Hongbo Fu}, {and} \bibinfo{person}{Lin Gao}.}
  \bibinfo{year}{2023}\natexlab{}.
\newblock \showarticletitle{NeRFFaceLighting: Implicit and Disentangled Face
  Lighting Representation Leveraging Generative Prior in Neural Radiance
  Fields}.
\newblock \bibinfo{journal}{\emph{ACM Transactions on Graphics (TOG)}}
  (\bibinfo{year}{2023}).
\newblock


\bibitem[Karras et~al\mbox{.}(2018)]%
        {Karras18}
\bibfield{author}{\bibinfo{person}{Tero Karras}, \bibinfo{person}{Timo Aila},
  \bibinfo{person}{Samuli Laine}, {and} \bibinfo{person}{Jaakko Lehtinen}.}
  \bibinfo{year}{2018}\natexlab{}.
\newblock \showarticletitle{Progressive Growing of GANs for Improved Quality,
  Stability, and Variation}. In \bibinfo{booktitle}{\emph{International
  Conference on Learning Representations (ICLR)}}.
\newblock


\bibitem[Karras et~al\mbox{.}(2020)]%
        {Karras2019stylegan2}
\bibfield{author}{\bibinfo{person}{Tero Karras}, \bibinfo{person}{Samuli
  Laine}, \bibinfo{person}{Miika Aittala}, \bibinfo{person}{Janne Hellsten},
  \bibinfo{person}{Jaakko Lehtinen}, {and} \bibinfo{person}{Timo Aila}.}
  \bibinfo{year}{2020}\natexlab{}.
\newblock \showarticletitle{Analyzing and Improving the Image Quality of
  {StyleGAN}}. In \bibinfo{booktitle}{\emph{Proc. CVPR}}.
\newblock


\bibitem[Kingma and Ba(2015)]%
        {KingmaB14}
\bibfield{author}{\bibinfo{person}{Diederik~P. Kingma} {and}
  \bibinfo{person}{Jimmy Ba}.} \bibinfo{year}{2015}\natexlab{}.
\newblock \showarticletitle{Adam: {A} Method for Stochastic Optimization}. In
  \bibinfo{booktitle}{\emph{3rd International Conference on Learning
  Representations, {ICLR} 2015, San Diego, CA, USA, May 7-9, 2015, Conference
  Track Proceedings}}, \bibfield{editor}{\bibinfo{person}{Yoshua Bengio} {and}
  \bibinfo{person}{Yann LeCun}} (Eds.).
\newblock
\urldef\tempurl%
\url{http://arxiv.org/abs/1412.6980}
\showURL{%
\tempurl}


\bibitem[Klehm et~al\mbox{.}(2015)]%
        {klehm15star}
\bibfield{author}{\bibinfo{person}{Oliver Klehm}, \bibinfo{person}{Fabrice
  Rousselle}, \bibinfo{person}{Marios Papas}, \bibinfo{person}{Derek Bradley},
  \bibinfo{person}{Christophe Hery}, \bibinfo{person}{Bernd Bickel},
  \bibinfo{person}{Wojciech Jarosz}, {and} \bibinfo{person}{Thabo Beeler}.}
  \bibinfo{year}{2015}\natexlab{}.
\newblock \showarticletitle{Recent Advances in Facial Appearance Capture}.
\newblock \bibinfo{journal}{\emph{Computer Graphics Forum (Proceedings of
  Eurographics - State of the Art Reports)}} \bibinfo{volume}{34},
  \bibinfo{number}{2} (\bibinfo{date}{May} \bibinfo{year}{2015}),
  \bibinfo{pages}{709–733}.
\newblock
\urldef\tempurl%
\url{https://doi.org/10/f7mb4b}
\showDOI{\tempurl}


\bibitem[Kwak et~al\mbox{.}(2022)]%
        {kwak2022injecting}
\bibfield{author}{\bibinfo{person}{Jeong-gi Kwak}, \bibinfo{person}{Yuanming
  Li}, \bibinfo{person}{Dongsik Yoon}, \bibinfo{person}{Donghyeon Kim},
  \bibinfo{person}{David Han}, {and} \bibinfo{person}{Hanseok Ko}.}
  \bibinfo{year}{2022}\natexlab{}.
\newblock \showarticletitle{Injecting 3D Perception of Controllable NeRF-GAN
  into StyleGAN for Editable Portrait Image Synthesis}. In
  \bibinfo{booktitle}{\emph{European Conference on Computer Vision}}. Springer,
  \bibinfo{pages}{236--253}.
\newblock


\bibitem[Lattas et~al\mbox{.}(2021)]%
        {lattas2021avatarme++}
\bibfield{author}{\bibinfo{person}{Alexandros Lattas},
  \bibinfo{person}{Stylianos Moschoglou}, \bibinfo{person}{Stylianos Ploumpis},
  \bibinfo{person}{Baris Gecer}, \bibinfo{person}{Abhijeet Ghosh}, {and}
  \bibinfo{person}{Stefanos~P Zafeiriou}.} \bibinfo{year}{2021}\natexlab{}.
\newblock \showarticletitle{AvatarMe++: Facial Shape and BRDF Inference with
  Photorealistic Rendering-Aware GANs}.
\newblock \bibinfo{journal}{\emph{IEEE Transactions on Pattern Analysis and
  Machine Intelligence}} (\bibinfo{year}{2021}).
\newblock


\bibitem[Li et~al\mbox{.}(2023)]%
        {megane_li_cvpr}
\bibfield{author}{\bibinfo{person}{Junxuan Li}, \bibinfo{person}{Shunsuke
  Saito}, \bibinfo{person}{Tomas Simon}, \bibinfo{person}{Stephen Lombardi},
  \bibinfo{person}{Hongdong Li}, {and} \bibinfo{person}{Jason Saragih}.}
  \bibinfo{year}{2023}\natexlab{}.
\newblock \showarticletitle{MEGANE: Morphable Eyeglass and Avatar Network}. In
  \bibinfo{booktitle}{\emph{Proceedings of the IEEE/CVF Conference on Computer
  Vision and Pattern Recognition (CVPR)}}. \bibinfo{pages}{12769--12779}.
\newblock


\bibitem[Lin et~al\mbox{.}(2020)]%
        {BGMv2}
\bibfield{author}{\bibinfo{person}{Shanchuan Lin}, \bibinfo{person}{Andrey
  Ryabtsev}, \bibinfo{person}{Soumyadip Sengupta}, \bibinfo{person}{Brian
  Curless}, \bibinfo{person}{Steve Seitz}, {and} \bibinfo{person}{Ira
  Kemelmacher-Shlizerman}.} \bibinfo{year}{2020}\natexlab{}.
\newblock \showarticletitle{Real-Time High-Resolution Background Matting}.
\newblock \bibinfo{journal}{\emph{arXiv}} (\bibinfo{year}{2020}),
  \bibinfo{pages}{arXiv--2012}.
\newblock


\bibitem[Livingstone and Russo(2018)]%
        {ravdess}
\bibfield{author}{\bibinfo{person}{Steven~R. Livingstone} {and}
  \bibinfo{person}{Frank~A. Russo}.} \bibinfo{year}{2018}\natexlab{}.
\newblock \showarticletitle{The Ryerson Audio-Visual Database of Emotional
  Speech and Song (RAVDESS): A dynamic, multimodal set of facial and vocal
  expressions in North American English}.
\newblock \bibinfo{journal}{\emph{PLOS ONE}} \bibinfo{volume}{13},
  \bibinfo{number}{5} (\bibinfo{date}{05} \bibinfo{year}{2018}),
  \bibinfo{pages}{1--35}.
\newblock
\urldef\tempurl%
\url{https://doi.org/10.1371/journal.pone.0196391}
\showDOI{\tempurl}


\bibitem[Martinez et~al\mbox{.}(2024)]%
        {martinez2024codec}
\bibfield{author}{\bibinfo{person}{Julieta Martinez}, \bibinfo{person}{Emily
  Kim}, \bibinfo{person}{Javier Romero}, \bibinfo{person}{Timur Bagautdinov},
  \bibinfo{person}{Shunsuke Saito}, \bibinfo{person}{Shoou-I Yu},
  \bibinfo{person}{Stuart Anderson}, \bibinfo{person}{Michael Zollhöfer},
  \bibinfo{person}{Te-Li Wang}, \bibinfo{person}{Shaojie Bai},
  \bibinfo{person}{Chenghui Li}, \bibinfo{person}{Shih-En Wei},
  \bibinfo{person}{Rohan Joshi}, \bibinfo{person}{Wyatt Borsos},
  \bibinfo{person}{Tomas Simon}, \bibinfo{person}{Jason Saragih},
  \bibinfo{person}{Paul Theodosis}, \bibinfo{person}{Alexander Greene},
  \bibinfo{person}{Anjani Josyula}, \bibinfo{person}{Silvio~Mano Maeta},
  \bibinfo{person}{Andrew~I. Jewett}, \bibinfo{person}{Simon Venshtain},
  \bibinfo{person}{Christopher Heilman}, \bibinfo{person}{Yueh-Tung Chen},
  \bibinfo{person}{Sidi Fu}, \bibinfo{person}{Mohamed Ezzeldin~A. Elshaer},
  \bibinfo{person}{Tingfang Du}, \bibinfo{person}{Longhua Wu},
  \bibinfo{person}{Shen-Chi Chen}, \bibinfo{person}{Kai Kang},
  \bibinfo{person}{Michael Wu}, \bibinfo{person}{Youssef Emad},
  \bibinfo{person}{Steven Longay}, \bibinfo{person}{Ashley Brewer},
  \bibinfo{person}{Hitesh Shah}, \bibinfo{person}{James Booth},
  \bibinfo{person}{Taylor Koska}, \bibinfo{person}{Kayla Haidle},
  \bibinfo{person}{Matt Andromalos}, \bibinfo{person}{Joanna Hsu},
  \bibinfo{person}{Thomas Dauer}, \bibinfo{person}{Peter Selednik},
  \bibinfo{person}{Tim Godisart}, \bibinfo{person}{Scott Ardisson},
  \bibinfo{person}{Matthew Cipperly}, \bibinfo{person}{Ben Humberston},
  \bibinfo{person}{Lon Farr}, \bibinfo{person}{Bob Hansen},
  \bibinfo{person}{Peihong Guo}, \bibinfo{person}{Dave Braun},
  \bibinfo{person}{Steven Krenn}, \bibinfo{person}{He Wen},
  \bibinfo{person}{Lucas Evans}, \bibinfo{person}{Natalia Fadeeva},
  \bibinfo{person}{Matthew Stewart}, \bibinfo{person}{Gabriel Schwartz},
  \bibinfo{person}{Divam Gupta}, \bibinfo{person}{Gyeongsik Moon},
  \bibinfo{person}{Kaiwen Guo}, \bibinfo{person}{Yuan Dong},
  \bibinfo{person}{Yichen Xu}, \bibinfo{person}{Takaaki Shiratori},
  \bibinfo{person}{Fabian Prada}, \bibinfo{person}{Bernardo~R. Pires},
  \bibinfo{person}{Bo Peng}, \bibinfo{person}{Julia Buffalini},
  \bibinfo{person}{Autumn Trimble}, \bibinfo{person}{Kevyn McPhail},
  \bibinfo{person}{Melissa Schoeller}, {and} \bibinfo{person}{Yaser Sheikh}.}
  \bibinfo{year}{2024}\natexlab{}.
\newblock \showarticletitle{{Codec Avatar Studio: Paired Human Captures for
  Complete, Driveable, and Generalizable Avatars}}.
\newblock \bibinfo{journal}{\emph{NeurIPS Track on Datasets and Benchmarks}}
  (\bibinfo{year}{2024}).
\newblock


\bibitem[Mei et~al\mbox{.}(2024)]%
        {mei2024holo}
\bibfield{author}{\bibinfo{person}{Yiqun Mei}, \bibinfo{person}{Yu Zeng},
  \bibinfo{person}{He Zhang}, \bibinfo{person}{Zhixin Shu},
  \bibinfo{person}{Xuaner Zhang}, \bibinfo{person}{Sai Bi},
  \bibinfo{person}{Jianming Zhang}, \bibinfo{person}{HyunJoon Jung}, {and}
  \bibinfo{person}{Vishal~M Patel}.} \bibinfo{year}{2024}\natexlab{}.
\newblock \showarticletitle{Holo-Relighting: Controllable Volumetric Portrait
  Relighting from a Single Image}.
\newblock \bibinfo{journal}{\emph{arXiv preprint arXiv:2403.09632}}
  (\bibinfo{year}{2024}).
\newblock


\bibitem[Meka et~al\mbox{.}(2019)]%
        {Meka19}
\bibfield{author}{\bibinfo{person}{Abhimitra Meka}, \bibinfo{person}{Christian
  H\"{a}ne}, \bibinfo{person}{Rohit Pandey}, \bibinfo{person}{Michael
  Zollh\"{o}fer}, \bibinfo{person}{Sean Fanello}, \bibinfo{person}{Graham
  Fyffe}, \bibinfo{person}{Adarsh Kowdle}, \bibinfo{person}{Xueming Yu},
  \bibinfo{person}{Jay Busch}, \bibinfo{person}{Jason Dourgarian},
  \bibinfo{person}{Peter Denny}, \bibinfo{person}{Sofien Bouaziz},
  \bibinfo{person}{Peter Lincoln}, \bibinfo{person}{Matt Whalen},
  \bibinfo{person}{Geoff Harvey}, \bibinfo{person}{Jonathan Taylor},
  \bibinfo{person}{Shahram Izadi}, \bibinfo{person}{Andrea Tagliasacchi},
  \bibinfo{person}{Paul Debevec}, \bibinfo{person}{Christian Theobalt},
  \bibinfo{person}{Julien Valentin}, {and} \bibinfo{person}{Christoph
  Rhemann}.} \bibinfo{year}{2019}\natexlab{}.
\newblock \showarticletitle{Deep Reflectance Fields: High-Quality Facial
  Reflectance Field Inference from Color Gradient Illumination}.
\newblock \bibinfo{journal}{\emph{ACM Transactions on Graphics (Proceedings of
  SIGGRAPH)}} (\bibinfo{year}{2019}).
\newblock


\bibitem[Meng et~al\mbox{.}(2021)]%
        {meng2021magface_ID_loss}
\bibfield{author}{\bibinfo{person}{Qiang Meng}, \bibinfo{person}{Shichao Zhao},
  \bibinfo{person}{Zhida Huang}, {and} \bibinfo{person}{Feng Zhou}.}
  \bibinfo{year}{2021}\natexlab{}.
\newblock \showarticletitle{{MagFace}: A universal representation for face
  recognition and quality assessment}. In \bibinfo{booktitle}{\emph{CVPR}}.
\newblock


\bibitem[Mildenhall et~al\mbox{.}(2020)]%
        {mildenhall2020nerf}
\bibfield{author}{\bibinfo{person}{Ben Mildenhall}, \bibinfo{person}{Pratul~P.
  Srinivasan}, \bibinfo{person}{Matthew Tancik}, \bibinfo{person}{Jonathan~T.
  Barron}, \bibinfo{person}{Ravi Ramamoorthi}, {and} \bibinfo{person}{Ren Ng}.}
  \bibinfo{year}{2020}\natexlab{}.
\newblock \showarticletitle{NeRF: Representing Scenes as Neural Radiance Fields
  for View Synthesis}. In \bibinfo{booktitle}{\emph{European Conference on
  Computer Vision (ECCV)}}.
\newblock


\bibitem[Nestmeyer et~al\mbox{.}(2020)]%
        {nestmeyer2020faceRelighting}
\bibfield{author}{\bibinfo{person}{Thomas Nestmeyer},
  \bibinfo{person}{Jean-François Lalonde}, \bibinfo{person}{Iain Matthews},
  {and} \bibinfo{person}{Andreas~M Lehrmann}.} \bibinfo{year}{2020}\natexlab{}.
\newblock \showarticletitle{Learning Physics-guided Face Relighting under
  Directional Light}. In \bibinfo{booktitle}{\emph{The IEEE Conference on
  Computer Vision and Pattern Recognition (CVPR)}}.
\newblock


\bibitem[Pandey et~al\mbox{.}(2021)]%
        {Pandey21}
\bibfield{author}{\bibinfo{person}{Rohit Pandey}, \bibinfo{person}{Sergio
  Orts-Escolano}, \bibinfo{person}{Chloe LeGendre}, \bibinfo{person}{Christian
  Haene}, \bibinfo{person}{Sofien Bouaziz}, \bibinfo{person}{Christoph
  Rhemann}, \bibinfo{person}{Paul Debevec}, {and} \bibinfo{person}{Sean
  Fanello}.} \bibinfo{year}{2021}\natexlab{}.
\newblock \showarticletitle{Total Relighting: Learning to Relight Portraits for
  Background Replacement}.
\newblock \bibinfo{journal}{\emph{ACM Transactions on Graphics (Proceedings
  SIGGRAPH)}} (\bibinfo{year}{2021}).
\newblock


\bibitem[Ranjan et~al\mbox{.}(2023)]%
        {neural-3d-relightable}
\bibfield{author}{\bibinfo{person}{Anurag Ranjan}, \bibinfo{person}{Kwang~Moo
  Yi}, \bibinfo{person}{Jen-Hao~Rick Chang}, {and} \bibinfo{person}{Oncel
  Tuzel}.} \bibinfo{year}{2023}\natexlab{}.
\newblock \showarticletitle{FaceLit: Neural 3D Relightable Faces}. In
  \bibinfo{booktitle}{\emph{CVPR}}.
\newblock
\urldef\tempurl%
\url{https://arxiv.org/abs/2303.15437}
\showURL{%
\tempurl}


\bibitem[Rao et~al\mbox{.}(2023)]%
        {prao2023vorf}
\bibfield{author}{\bibinfo{person}{Pramod Rao}, \bibinfo{person}{Mallikarjun
  B.~R}, \bibinfo{person}{Gereon Fox}, \bibinfo{person}{Tim Weyrich},
  \bibinfo{person}{Bernd Bickel}, \bibinfo{person}{Hanspeter Pfister},
  \bibinfo{person}{Wojciech Matusik}, \bibinfo{person}{Fangneng Zhan},
  \bibinfo{person}{Ayush Tewari}, \bibinfo{person}{Christian Theobalt}, {and}
  \bibinfo{person}{Elgharib Mohamed}.} \bibinfo{year}{2023}\natexlab{}.
\newblock \showarticletitle{A Deeper Analysis of Volumetric Relightable Faces}.
\newblock \bibinfo{journal}{\emph{International Journal of Computer Vision}}
  (\bibinfo{date}{10} \bibinfo{year}{2023}), \bibinfo{pages}{1--19}.
\newblock
\urldef\tempurl%
\url{https://doi.org/10.1007/s11263-023-01899-3}
\showDOI{\tempurl}


\bibitem[Rao et~al\mbox{.}(2022)]%
        {prao2022vorf}
\bibfield{author}{\bibinfo{person}{Pramod Rao}, \bibinfo{person}{Mallikarjun
  B~R}, \bibinfo{person}{Gereon Fox}, \bibinfo{person}{Tim Weyrich},
  \bibinfo{person}{Bernd Bickel}, \bibinfo{person}{Hans-Peter Seidel},
  \bibinfo{person}{Hanspeter Pfister}, \bibinfo{person}{Wojciech Matusik},
  \bibinfo{person}{Ayush Tewari}, \bibinfo{person}{Christian Theobalt}, {and}
  \bibinfo{person}{Mohamed Elgharib}.} \bibinfo{year}{2022}\natexlab{}.
\newblock \showarticletitle{VoRF: Volumetric Relightable Faces}.
\newblock  (\bibinfo{year}{2022}).
\newblock


\bibitem[Rao et~al\mbox{.}(2024a)]%
        {prao2024lite2relight}
\bibfield{author}{\bibinfo{person}{Pramod Rao}, \bibinfo{person}{Gereon Fox},
  \bibinfo{person}{Abhimitra Meka}, \bibinfo{person}{Mallikarjun B~R},
  \bibinfo{person}{Fangneng Zhan}, \bibinfo{person}{Tim Weyrich},
  \bibinfo{person}{Bernd Bickel}, \bibinfo{person}{Hans-Peter Seidel},
  \bibinfo{person}{Hanspeter Pfister}, \bibinfo{person}{Wojciech Matusik},
  \bibinfo{person}{Mohamed Elgharib}, {and} \bibinfo{person}{Christian
  Theobalt}.} \bibinfo{year}{2024}\natexlab{a}.
\newblock \showarticletitle{Lite2Relight: 3D-aware Single Image Portrait
  Relighting}.
\newblock  (\bibinfo{year}{2024}).
\newblock


\bibitem[Rao et~al\mbox{.}(2024b)]%
        {prao2024l2r}
\bibfield{author}{\bibinfo{person}{Pramod Rao}, \bibinfo{person}{Gereon Fox},
  \bibinfo{person}{Abhimitra Meka}, \bibinfo{person}{Mallikarjun B~R},
  \bibinfo{person}{Fangneng Zhan}, \bibinfo{person}{Tim Weyrich},
  \bibinfo{person}{Bernd Bickel}, \bibinfo{person}{Hans-Peter Seidel},
  \bibinfo{person}{Hanspeter Pfister}, \bibinfo{person}{Wojciech Matusik},
  \bibinfo{person}{Mohamed Elgharib}, {and} \bibinfo{person}{Christian
  Theobalt}.} \bibinfo{year}{2024}\natexlab{b}.
\newblock \showarticletitle{Lite2Relight: 3D-aware Single Image Portrait
  Relighting}.
\newblock  (\bibinfo{year}{2024}).
\newblock


\bibitem[Rudnev et~al\mbox{.}(2022)]%
        {rudnev2022nerfosr}
\bibfield{author}{\bibinfo{person}{Viktor Rudnev}, \bibinfo{person}{Mohamed
  Elgharib}, \bibinfo{person}{William Smith}, \bibinfo{person}{Lingjie Liu},
  \bibinfo{person}{Vladislav Golyanik}, {and} \bibinfo{person}{Christian
  Theobalt}.} \bibinfo{year}{2022}\natexlab{}.
\newblock \showarticletitle{NeRF for Outdoor Scene Relighting}. In
  \bibinfo{booktitle}{\emph{European Conference on Computer Vision (ECCV)}}.
\newblock


\bibitem[Saito et~al\mbox{.}(2024)]%
        {saito2024rgca}
\bibfield{author}{\bibinfo{person}{Shunsuke Saito}, \bibinfo{person}{Gabriel
  Schwartz}, \bibinfo{person}{Tomas Simon}, \bibinfo{person}{Junxuan Li}, {and}
  \bibinfo{person}{Giljoo Nam}.} \bibinfo{year}{2024}\natexlab{}.
\newblock \showarticletitle{Relightable Gaussian Codec Avatars}. In
  \bibinfo{booktitle}{\emph{CVPR}}.
\newblock


\bibitem[Sarkar et~al\mbox{.}(2023)]%
        {litnerf}
\bibfield{author}{\bibinfo{person}{Kripasindhu Sarkar},
  \bibinfo{person}{Marcel~C. B\"{u}hler}, \bibinfo{person}{Gengyan Li},
  \bibinfo{person}{Daoye Wang}, \bibinfo{person}{Delio Vicini},
  \bibinfo{person}{J\'{e}r\'{e}my Riviere}, \bibinfo{person}{Yinda Zhang},
  \bibinfo{person}{Sergio Orts-Escolano}, \bibinfo{person}{Paulo Gotardo},
  \bibinfo{person}{Thabo Beeler}, {and} \bibinfo{person}{Abhimitra Meka}.}
  \bibinfo{year}{2023}\natexlab{}.
\newblock \showarticletitle{LitNeRF: Intrinsic Radiance Decomposition for
  High-Quality View Synthesis and Relighting of Faces}. In
  \bibinfo{booktitle}{\emph{SIGGRAPH Asia 2023 Conference Papers}} (<conf-loc>,
  <city>Sydney</city>, <state>NSW</state>, <country>Australia</country>,
  </conf-loc>) \emph{(\bibinfo{series}{SA '23})}.
  \bibinfo{publisher}{Association for Computing Machinery},
  \bibinfo{address}{New York, NY, USA}, Article \bibinfo{articleno}{42},
  \bibinfo{numpages}{11}~pages.
\newblock
\showISBNx{9798400703157}
\urldef\tempurl%
\url{https://doi.org/10.1145/3610548.3618210}
\showDOI{\tempurl}


\bibitem[Sengupta et~al\mbox{.}(2018)]%
        {sfsnetSengupta18}
\bibfield{author}{\bibinfo{person}{Soumyadip Sengupta}, \bibinfo{person}{Angjoo
  Kanazawa}, \bibinfo{person}{Carlos~D. Castillo}, {and}
  \bibinfo{person}{David~W. Jacobs}.} \bibinfo{year}{2018}\natexlab{}.
\newblock \showarticletitle{SfSNet: Learning Shape, Refectance and Illuminance
  of Faces in the Wild}. In \bibinfo{booktitle}{\emph{The IEEE Conference on
  Computer Vision and Pattern Recognition (CVPR)}}.
\newblock


\bibitem[Shih et~al\mbox{.}(2014)]%
        {shih2014style}
\bibfield{author}{\bibinfo{person}{YiChang Shih}, \bibinfo{person}{Sylvain
  Paris}, \bibinfo{person}{Connelly Barnes}, \bibinfo{person}{William~T.
  Freeman}, {and} \bibinfo{person}{Fr\'{e}do Durand}.}
  \bibinfo{year}{2014}\natexlab{}.
\newblock \showarticletitle{Style transfer for headshot portraits}.
\newblock \bibinfo{journal}{\emph{ACM Trans. Graph.}} \bibinfo{volume}{33},
  \bibinfo{number}{4}, Article \bibinfo{articleno}{148} (\bibinfo{date}{July}
  \bibinfo{year}{2014}), \bibinfo{numpages}{14}~pages.
\newblock
\showISSN{0730-0301}
\urldef\tempurl%
\url{https://doi.org/10.1145/2601097.2601137}
\showDOI{\tempurl}


\bibitem[{Shu} et~al\mbox{.}(2017)]%
        {Shu17NeuralEditing}
\bibfield{author}{\bibinfo{person}{Z. {Shu}}, \bibinfo{person}{E. {Yumer}},
  \bibinfo{person}{S. {Hadap}}, \bibinfo{person}{K. {Sunkavalli}},
  \bibinfo{person}{E. {Shechtman}}, {and} \bibinfo{person}{D. {Samaras}}.}
  \bibinfo{year}{2017}\natexlab{}.
\newblock \showarticletitle{Neural Face Editing with Intrinsic Image
  Disentangling}. In \bibinfo{booktitle}{\emph{The IEEE Conference on Computer
  Vision and Pattern Recognition (CVPR)}}.
\newblock


\bibitem[Simonyan and Zisserman(2015)]%
        {Simonyan15}
\bibfield{author}{\bibinfo{person}{Karen Simonyan} {and}
  \bibinfo{person}{Andrew Zisserman}.} \bibinfo{year}{2015}\natexlab{}.
\newblock \showarticletitle{Very Deep Convolutional Networks for Large-Scale
  Image Recognition}. In \bibinfo{booktitle}{\emph{International Conference on
  Learning Representations}}.
\newblock


\bibitem[Sklyarova et~al\mbox{.}(2023)]%
        {hair_sklyarova2023haar}
\bibfield{author}{\bibinfo{person}{Vanessa Sklyarova}, \bibinfo{person}{Egor
  Zakharov}, \bibinfo{person}{Otmar Hilliges}, \bibinfo{person}{Michael~J
  Black}, {and} \bibinfo{person}{Justus Thies}.}
  \bibinfo{year}{2023}\natexlab{}.
\newblock \showarticletitle{HAAR: Text-Conditioned Generative Model of 3D
  Strand-based Human Hairstyles}.
\newblock \bibinfo{journal}{\emph{ArXiv}} (\bibinfo{date}{Dec}
  \bibinfo{year}{2023}).
\newblock


\bibitem[Srinivasan et~al\mbox{.}(2021)]%
        {Srinivasan21NeRV}
\bibfield{author}{\bibinfo{person}{Pratul~P. Srinivasan},
  \bibinfo{person}{Boyang Deng}, \bibinfo{person}{Xiuming Zhang},
  \bibinfo{person}{Matthew Tancik}, \bibinfo{person}{Ben Mildenhall}, {and}
  \bibinfo{person}{Jonathan~T. Barron}.} \bibinfo{year}{2021}\natexlab{}.
\newblock \showarticletitle{NeRV: Neural Reflectance and Visibility Fields for
  Relighting and View Synthesis}. In \bibinfo{booktitle}{\emph{The IEEE
  Conference on Computer Vision and Pattern Recognition (CVPR)}}.
\newblock


\bibitem[Stratou et~al\mbox{.}(2011)]%
        {stratou2011ICTFaceDB}
\bibfield{author}{\bibinfo{person}{Giota Stratou}, \bibinfo{person}{Abhijeet
  Ghosh}, \bibinfo{person}{Paul Debevec}, {and} \bibinfo{person}{Louis-Philippe
  Morency}.} \bibinfo{year}{2011}\natexlab{}.
\newblock \showarticletitle{Effect of illumination on automatic expression
  recognition: A novel 3D relightable facial database}. In
  \bibinfo{booktitle}{\emph{2011 IEEE International Conference on Automatic
  Face \& Gesture Recognition (FG)}}. \bibinfo{pages}{611--618}.
\newblock
\urldef\tempurl%
\url{https://doi.org/10.1109/FG.2011.5771467}
\showDOI{\tempurl}


\bibitem[Sun et~al\mbox{.}(2021)]%
        {sun2021nelf}
\bibfield{author}{\bibinfo{person}{Tiancheng Sun}, \bibinfo{person}{Kai-En
  Lin}, \bibinfo{person}{Sai Bi}, \bibinfo{person}{Zexiang Xu}, {and}
  \bibinfo{person}{Ravi Ramamoorthi}.} \bibinfo{year}{2021}\natexlab{}.
\newblock \showarticletitle{{NeLF}: Neural Light-transport Field for Portrait
  View Synthesis and Relighting}. In \bibinfo{booktitle}{\emph{Eurographics
  Symposium on Rendering}}.
\newblock


\bibitem[Sun et~al\mbox{.}(2020)]%
        {sun2020light}
\bibfield{author}{\bibinfo{person}{Tiancheng Sun}, \bibinfo{person}{Zexiang
  Xu}, \bibinfo{person}{Xiuming Zhang}, \bibinfo{person}{Sean Fanello},
  \bibinfo{person}{Christoph Rhemann}, \bibinfo{person}{Paul Debevec},
  \bibinfo{person}{Yun-Ta Tsai}, \bibinfo{person}{Jonathan~T. Barron}, {and}
  \bibinfo{person}{Ravi Ramamoorthi}.} \bibinfo{year}{2020}\natexlab{}.
\newblock \showarticletitle{Light Stage Super-Resolution: Continuous
  High-Frequency Relighting}. In \bibinfo{booktitle}{\emph{ACM Transactions on
  Graphics (Proceedings of SIGGRAPH Asia)}}.
\newblock


\bibitem[Tan et~al\mbox{.}(2022)]%
        {tan2022voluxgan}
\bibfield{author}{\bibinfo{person}{Feitong Tan}, \bibinfo{person}{Sean
  Fanello}, \bibinfo{person}{Abhimitra Meka}, \bibinfo{person}{Sergio
  Orts-Escolano}, \bibinfo{person}{Danhang Tang}, \bibinfo{person}{Rohit
  Pandey}, \bibinfo{person}{Jonathan Taylor}, \bibinfo{person}{Ping Tan}, {and}
  \bibinfo{person}{Yinda Zhang}.} \bibinfo{year}{2022}\natexlab{}.
\newblock \bibinfo{title}{VoLux-GAN: A Generative Model for 3D Face Synthesis
  with HDRI Relighting}.
\newblock
\newblock
\showeprint[arxiv]{2201.04873}~[cs.CV]


\bibitem[Teed and Deng(2020)]%
        {Teed2020RAFT}
\bibfield{author}{\bibinfo{person}{Zachary Teed} {and} \bibinfo{person}{Jia
  Deng}.} \bibinfo{year}{2020}\natexlab{}.
\newblock \showarticletitle{RAFT: Recurrent All-Pairs Field Transforms for
  Optical Flow}. In \bibinfo{booktitle}{\emph{European Conference on Computer
  Vision}}.
\newblock


\bibitem[Tewari et~al\mbox{.}(2020a)]%
        {stylerig}
\bibfield{author}{\bibinfo{person}{Ayush Tewari}, \bibinfo{person}{Mohamed
  Elgharib}, \bibinfo{person}{Gaurav Bharaj}, \bibinfo{person}{Florian
  Bernard}, \bibinfo{person}{Hans-Peter Seidel}, \bibinfo{person}{Patrick
  P{\'e}rez}, \bibinfo{person}{Michael Z{\"o}llhofer}, {and}
  \bibinfo{person}{Christian Theobalt}.} \bibinfo{year}{2020}\natexlab{a}.
\newblock \showarticletitle{StyleRig: Rigging StyleGAN for 3D Control over
  Portrait Images, CVPR 2020}. In \bibinfo{booktitle}{\emph{{IEEE} Conference
  on Computer Vision and Pattern Recognition (CVPR)}}. {IEEE}.
\newblock


\bibitem[Tewari et~al\mbox{.}(2020b)]%
        {tewari2020pie}
\bibfield{author}{\bibinfo{person}{Ayush Tewari}, \bibinfo{person}{Mohamed
  Elgharib}, \bibinfo{person}{Mallikarjun BR}, \bibinfo{person}{Florian
  Bernard}, \bibinfo{person}{Hans-Peter Seidel}, \bibinfo{person}{Patrick
  P{\'e}rez}, \bibinfo{person}{Michael Z{\"o}llhofer}, {and}
  \bibinfo{person}{Christian Theobalt}.} \bibinfo{year}{2020}\natexlab{b}.
\newblock \showarticletitle{PIE: Portrait Image Embedding for Semantic
  Control}.
\newblock \bibinfo{journal}{\emph{ACM Transactions on Graphics (Proceedings
  SIGGRAPH Asia)}} \bibinfo{volume}{39}, \bibinfo{number}{6}
  (\bibinfo{date}{December} \bibinfo{year}{2020}).
\newblock
\urldef\tempurl%
\url{https://doi.org/10.1145/3414685.3417803}
\showDOI{\tempurl}


\bibitem[Wang et~al\mbox{.}(2018)]%
        {wang2018mrfloss}
\bibfield{author}{\bibinfo{person}{Yi Wang}, \bibinfo{person}{Xin Tao},
  \bibinfo{person}{Xiaojuan Qi}, \bibinfo{person}{Xiaoyong Shen}, {and}
  \bibinfo{person}{Jiaya Jia}.} \bibinfo{year}{2018}\natexlab{}.
\newblock \showarticletitle{Image Inpainting via Generative Multi-column
  Convolutional Neural Networks}. In \bibinfo{booktitle}{\emph{Advances in
  Neural Information Processing Systems}}. \bibinfo{pages}{331--340}.
\newblock


\bibitem[Wang et~al\mbox{.}(2020)]%
        {sipr_ex}
\bibfield{author}{\bibinfo{person}{Zhibo Wang}, \bibinfo{person}{Xin Yu},
  \bibinfo{person}{Ming Lu}, \bibinfo{person}{Quan Wang}, \bibinfo{person}{Chen
  Qian}, {and} \bibinfo{person}{Feng Xu}.} \bibinfo{year}{2020}\natexlab{}.
\newblock \showarticletitle{Single Image Portrait Relighting via Explicit
  Multiple Reflectance Channel Modeling}.
\newblock \bibinfo{journal}{\emph{ACM Transactions on Graphics (Proceedings of
  SIGGRAPH Asia)}} (\bibinfo{year}{2020}).
\newblock


\bibitem[Wenger et~al\mbox{.}(2005)]%
        {wenger_2005_sigg}
\bibfield{author}{\bibinfo{person}{Andreas Wenger}, \bibinfo{person}{Andrew
  Gardner}, \bibinfo{person}{Chris Tchou}, \bibinfo{person}{Jonas Unger},
  \bibinfo{person}{Tim Hawkins}, {and} \bibinfo{person}{Paul Debevec}.}
  \bibinfo{year}{2005}\natexlab{}.
\newblock \showarticletitle{Performance relighting and reflectance
  transformation with time-multiplexed illumination}.
\newblock \bibinfo{journal}{\emph{ACM Trans. Graph.}} \bibinfo{volume}{24},
  \bibinfo{number}{3} (\bibinfo{date}{July} \bibinfo{year}{2005}),
  \bibinfo{pages}{756–764}.
\newblock
\showISSN{0730-0301}
\urldef\tempurl%
\url{https://doi.org/10.1145/1073204.1073258}
\showDOI{\tempurl}


\bibitem[Weyrich et~al\mbox{.}(2006)]%
        {Weyrich2006Analysis}
\bibfield{author}{\bibinfo{person}{Tim Weyrich}, \bibinfo{person}{Wojciech
  Matusik}, \bibinfo{person}{Hanspeter Pfister}, \bibinfo{person}{Bernd
  Bickel}, \bibinfo{person}{Craig Donner}, \bibinfo{person}{Chien Tu},
  \bibinfo{person}{Janet McAndless}, \bibinfo{person}{Jinho Lee},
  \bibinfo{person}{Addy Ngan}, \bibinfo{person}{Henrik~Wann Jensen}, {and}
  \bibinfo{person}{Markus Gross}.} \bibinfo{year}{2006}\natexlab{}.
\newblock \showarticletitle{Analysis of Human Faces using a Measurement-Based
  Skin Reflectance Model}.
\newblock \bibinfo{journal}{\emph{ACM Transactions on Graphics (Proceedings of
  SIGGRAPH)}} (\bibinfo{year}{2006}).
\newblock


\bibitem[Yamaguchi et~al\mbox{.}(2018)]%
        {yamaguchi18}
\bibfield{author}{\bibinfo{person}{Shuco Yamaguchi}, \bibinfo{person}{Shunsuke
  Saito}, \bibinfo{person}{Koki Nagano}, \bibinfo{person}{Yajie Zhao},
  \bibinfo{person}{Weikai Chen}, \bibinfo{person}{Kyle Olszewski},
  \bibinfo{person}{Shigeo Morishima}, {and} \bibinfo{person}{Hao Li}.}
  \bibinfo{year}{2018}\natexlab{}.
\newblock \showarticletitle{High-fidelity facial reflectance and geometry
  inference from an unconstrained image}.
\newblock \bibinfo{journal}{\emph{ACM Transactions on Graphics (Proceedings of
  SIGGRAPH)}} (\bibinfo{year}{2018}).
\newblock


\bibitem[Yang et~al\mbox{.}(2023)]%
        {yang2023towards}
\bibfield{author}{\bibinfo{person}{Haotian Yang}, \bibinfo{person}{Mingwu
  Zheng}, \bibinfo{person}{Wanquan Feng}, \bibinfo{person}{Haibin Huang},
  \bibinfo{person}{Yu-Kun Lai}, \bibinfo{person}{Pengfei Wan},
  \bibinfo{person}{Zhongyuan Wang}, {and} \bibinfo{person}{Chongyang Ma}.}
  \bibinfo{year}{2023}\natexlab{}.
\newblock \showarticletitle{Towards practical capture of high-fidelity
  relightable avatars}. In \bibinfo{booktitle}{\emph{SIGGRAPH Asia 2023
  Conference Papers}}. \bibinfo{pages}{1--11}.
\newblock


\bibitem[Yu et~al\mbox{.}(2021)]%
        {yu2020pixelnerf}
\bibfield{author}{\bibinfo{person}{Alex Yu}, \bibinfo{person}{Vickie Ye},
  \bibinfo{person}{Matthew Tancik}, {and} \bibinfo{person}{Angjoo Kanazawa}.}
  \bibinfo{year}{2021}\natexlab{}.
\newblock \showarticletitle{{pixelNeRF}: Neural Radiance Fields from One or Few
  Images}. In \bibinfo{booktitle}{\emph{CVPR}}.
\newblock


\bibitem[Yuan et~al\mbox{.}(2023)]%
        {goae}
\bibfield{author}{\bibinfo{person}{Ziyang Yuan}, \bibinfo{person}{Yiming Zhu},
  \bibinfo{person}{Yu Li}, \bibinfo{person}{Hongyu Liu}, {and}
  \bibinfo{person}{Chun Yuan}.} \bibinfo{year}{2023}\natexlab{}.
\newblock \showarticletitle{Make Encoder Great Again in 3D GAN Inversion
  through Geometry and Occlusion-Aware Encoding}.
\newblock \bibinfo{journal}{\emph{arXiv preprint arXiv:2303.12326}}
  (\bibinfo{year}{2023}).
\newblock


\bibitem[Zakharov et~al\mbox{.}(2024)]%
        {hair_zakharov2024gh}
\bibfield{author}{\bibinfo{person}{Egor Zakharov}, \bibinfo{person}{Vanessa
  Sklyarova}, \bibinfo{person}{Michael~J Black}, \bibinfo{person}{Giljoo Nam},
  \bibinfo{person}{Justus Thies}, {and} \bibinfo{person}{Otmar Hilliges}.}
  \bibinfo{year}{2024}\natexlab{}.
\newblock \showarticletitle{Human Hair Reconstruction with Strand-Aligned 3D
  Gaussians}.
\newblock \bibinfo{journal}{\emph{ArXiv}} (\bibinfo{date}{Sep}
  \bibinfo{year}{2024}).
\newblock


\bibitem[Zeng et~al\mbox{.}(2024)]%
        {zeng2024_dilightnet}
\bibfield{author}{\bibinfo{person}{Chong Zeng}, \bibinfo{person}{Yue Dong},
  \bibinfo{person}{Pieter Peers}, \bibinfo{person}{Youkang Kong},
  \bibinfo{person}{Hongzhi Wu}, {and} \bibinfo{person}{Xin Tong}.}
  \bibinfo{year}{2024}\natexlab{}.
\newblock \showarticletitle{DiLightNet: Fine-grained Lighting Control for
  Diffusion-based Image Generation}. In \bibinfo{booktitle}{\emph{ACM SIGGRAPH
  2024 Conference Papers}}.
\newblock


\bibitem[Zhang et~al\mbox{.}(2025)]%
        {zhang2025_ic_light}
\bibfield{author}{\bibinfo{person}{Lvmin Zhang}, \bibinfo{person}{Anyi Rao},
  {and} \bibinfo{person}{Maneesh Agrawala}.} \bibinfo{year}{2025}\natexlab{}.
\newblock \showarticletitle{Scaling In-the-Wild Training for Diffusion-based
  Illumination Harmonization and Editing by Imposing Consistent Light
  Transport}. In \bibinfo{booktitle}{\emph{The Thirteenth International
  Conference on Learning Representations}}.
\newblock
\urldef\tempurl%
\url{https://openreview.net/forum?id=u1cQYxRI1H}
\showURL{%
\tempurl}


\bibitem[Zhang et~al\mbox{.}(2021b)]%
        {zhang2021dynamicolat}
\bibfield{author}{\bibinfo{person}{Longwen Zhang}, \bibinfo{person}{Qixuan
  Zhang}, \bibinfo{person}{Minye Wu}, \bibinfo{person}{Jingyi Yu}, {and}
  \bibinfo{person}{Lan Xu}.} \bibinfo{year}{2021}\natexlab{b}.
\newblock \showarticletitle{Neural Video Portrait Relighting in Real-Time via
  Consistency Modeling}. In \bibinfo{booktitle}{\emph{Proceedings of the
  IEEE/CVF International Conference on Computer Vision (ICCV)}}.
  \bibinfo{pages}{802--812}.
\newblock


\bibitem[Zhang et~al\mbox{.}(2018)]%
        {zhang2018perceptual}
\bibfield{author}{\bibinfo{person}{Richard Zhang}, \bibinfo{person}{Phillip
  Isola}, \bibinfo{person}{Alexei~A Efros}, \bibinfo{person}{Eli Shechtman},
  {and} \bibinfo{person}{Oliver Wang}.} \bibinfo{year}{2018}\natexlab{}.
\newblock \showarticletitle{The Unreasonable Effectiveness of Deep Features as
  a Perceptual Metric}. In \bibinfo{booktitle}{\emph{CVPR}}.
\newblock


\bibitem[Zhang et~al\mbox{.}(2020)]%
        {zhang2020portrait}
\bibfield{author}{\bibinfo{person}{Xuaner Zhang}, \bibinfo{person}{Jonathan~T.
  Barron}, \bibinfo{person}{Yun-Ta Tsai}, \bibinfo{person}{Rohit Pandey},
  \bibinfo{person}{Xiuming Zhang}, \bibinfo{person}{Ren Ng}, {and}
  \bibinfo{person}{David~E. Jacobs}.} \bibinfo{year}{2020}\natexlab{}.
\newblock \showarticletitle{Portrait Shadow Manipulation}. In
  \bibinfo{booktitle}{\emph{ACM Transactions on Graphics (TOG)}}.
\newblock


\bibitem[Zhang et~al\mbox{.}(2021a)]%
        {Zhang21}
\bibfield{author}{\bibinfo{person}{Xiuming Zhang}, \bibinfo{person}{Pratul~P.
  Srinivasan}, \bibinfo{person}{Boyang Deng}, \bibinfo{person}{Paul Debevec},
  \bibinfo{person}{William~T. Freeman}, {and} \bibinfo{person}{Jonathan~T.
  Barron}.} \bibinfo{year}{2021}\natexlab{a}.
\newblock \showarticletitle{NeRFactor: Neural Factorization of Shape and
  Reflectance under an Unknown Illumination}.
\newblock \bibinfo{journal}{\emph{ACM Transactions on Graphics}}
  (\bibinfo{year}{2021}).
\newblock


\bibitem[Zheng et~al\mbox{.}(2024)]%
        {rmbgv2_BiRefNet}
\bibfield{author}{\bibinfo{person}{Peng Zheng}, \bibinfo{person}{Dehong Gao},
  \bibinfo{person}{Deng-Ping Fan}, \bibinfo{person}{Li Liu},
  \bibinfo{person}{Jorma Laaksonen}, \bibinfo{person}{Wanli Ouyang}, {and}
  \bibinfo{person}{Nicu Sebe}.} \bibinfo{year}{2024}\natexlab{}.
\newblock \showarticletitle{Bilateral Reference for High-Resolution Dichotomous
  Image Segmentation}.
\newblock \bibinfo{journal}{\emph{CAAI Artificial Intelligence Research}}
  (\bibinfo{year}{2024}).
\newblock


\bibitem[Zheng et~al\mbox{.}(2025)]%
        {hair_Zheng2025GroomLight}
\bibfield{author}{\bibinfo{person}{Yang Zheng}, \bibinfo{person}{Menglei Chai},
  \bibinfo{person}{Delio Vicini}, \bibinfo{person}{Yuxiao Zhou},
  \bibinfo{person}{Yinghao Xu}, \bibinfo{person}{Leonidas Guibas},
  \bibinfo{person}{Gordon Wetzstein}, {and} \bibinfo{person}{Thabo Beeler}.}
  \bibinfo{year}{2025}\natexlab{}.
\newblock \showarticletitle{GroomLight: Hybrid Inverse Rendering for
  Relightable Human Hair Appearance Modeling}.
\newblock \bibinfo{journal}{\emph{arxiv}}.
\newblock


\bibitem[Zhou et~al\mbox{.}(2019)]%
        {Zhou_2019_ICCV}
\bibfield{author}{\bibinfo{person}{Hao Zhou}, \bibinfo{person}{Sunil Hadap},
  \bibinfo{person}{Kalyan Sunkavalli}, {and} \bibinfo{person}{David~W.
  Jacobs}.} \bibinfo{year}{2019}\natexlab{}.
\newblock \showarticletitle{Deep Single-Image Portrait Relighting}. In
  \bibinfo{booktitle}{\emph{The IEEE International Conference on Computer
  Vision (ICCV)}}.
\newblock


\bibitem[Zhou et~al\mbox{.}(2023)]%
        {zhou2023relightable}
\bibfield{author}{\bibinfo{person}{Taotao Zhou}, \bibinfo{person}{Kai He},
  \bibinfo{person}{Di Wu}, \bibinfo{person}{Teng Xu}, \bibinfo{person}{Qixuan
  Zhang}, \bibinfo{person}{Kuixiang Shao}, \bibinfo{person}{Wenzheng Chen},
  \bibinfo{person}{Lan Xu}, {and} \bibinfo{person}{Jingyi Yu}.}
  \bibinfo{year}{2023}\natexlab{}.
\newblock \showarticletitle{Relightable Neural Human Assets from Multi-view
  Gradient Illuminations}. In \bibinfo{booktitle}{\emph{Proceedings of the
  IEEE/CVF Conference on Computer Vision and Pattern Recognition}}.
  \bibinfo{pages}{4315--4327}.
\newblock


\end{thebibliography}

\end{document}